\documentclass{interact}

\usepackage{url}
\usepackage{booktabs}
\usepackage{multirow}
\usepackage{graphicx}
\usepackage{siunitx}
\usepackage{tabularx}
\usepackage{subfigure}
\usepackage{amsmath}

\newcommand{\argmax}{\operatornamewithlimits{argmax}}

\newcolumntype{C}{>{\centering\arraybackslash}X}
\newcolumntype{L}{>{\raggedright\arraybackslash}X}
\newcolumntype{R}{>{\raggedleft\arraybackslash}X}

\begin{document}

\title{Learning Geometric and Photometric Features from Panoramic LiDAR Scans for Outdoor Place Categorization}

\author{
    Kazuto~Nakashima$^{a}$$^{\ast}$\thanks{$^\ast$Corresponding author. Email: k\_nakashima@irvs.ait.kyushu-u.ac.jp},
    Hojung~Jung$^{a}$,
    Yuki~Oto$^{a}$,
    Yumi~Iwashita$^{b}$,
    Ryo~Kurazume$^{c}$
    and Oscar~Martinez~Mozos$^{d}$\\
    \vspace{6pt}
    $^{a}${\em{Graduate School of Information Science and Electrical Engineering, Kyushu University, Fukuoka, Japan}};
    $^{b}${\em{Jet Propulsion Laboratory, California Institute of Technology, Pasadena, CA, USA}};
    $^{c}${\em{Faculty of Information Science and Electrical Engineering, Kyushu University, Fukuoka, Japan}};
    $^{d}${\em{Technical University of Cartagena, Cartagena, Spain}}
}

\received{Published in Advanced Robotics on 31 Jul 2018}

\maketitle

\begin{abstract}
    Semantic place categorization, which is one of the essential tasks for autonomous robots and vehicles, allows them to have capabilities of self-decision and navigation in unfamiliar environments. In particular, outdoor places are more difficult targets than indoor ones due to perceptual variations, such as dynamic illuminance over twenty-four hours and occlusions by cars and pedestrians. This paper presents a novel method of categorizing outdoor places using convolutional neural networks (CNNs), which take omnidirectional depth/reflectance images obtained by 3D LiDARs as the inputs. First, we construct a large-scale outdoor place dataset named Multi-modal Panoramic 3D Outdoor (MPO) comprising two types of point clouds captured by two different LiDARs. They are labeled with six outdoor place categories: coast, forest, indoor/outdoor parking, residential area, and urban area. Second, we provide CNNs for LiDAR-based outdoor place categorization and evaluate our approach with the MPO dataset. Our results on the MPO dataset outperform traditional approaches and show the effectiveness in which we use both depth and reflectance modalities. To analyze our trained deep networks we visualize the learned features.
    \medskip
    \medskip
\end{abstract}
\section{Introduction}
Accurate and robust understanding of the surrounding environment is an important capability for autonomous robots and vehicles.
This understanding allows them to avoid an unexpected collision while moving and to make better decisions in context.
Meanwhile, due to dynamic changes and complexity of the real world, the task is challenging for robots and requires high-level interpretation ability of the observed sensor data and generalization capabilities to reason about a variety of environments including unseen previous ones.

In this paper, we address the problem of place recognition in which a robot needs to determine the type of place where it is located.
In particular, we focus on a categorization task that aims to predict the semantic place category.
Information about the place greatly improves communication between robots and humans~\cite{pronobis2012icra}. It also allows autonomous robots to make context-based decisions to complete high-level tasks~\cite{Stach2006, Koll2009}. Moreover, information about the type of the place can be used to build semantic maps of environments \cite{Prono2010} and high-level conceptual representations of a space~\cite{Chris2010}. Finally, an autonomous vehicle that is able to determine the type of its location can make better context-based decisions\cite{ibanez2012autonomous}. As an example, a vehicle can lower its speed when driving through a residential area.

As the most simple approach to determine the place category, geographic information measured using a global navigation satellite system (GNSS) is available.
Although robots can have a self-localization function immediately once they install light-weight dedicated sensors, the semantically labeled environmental map which should be extremely large and updated frequently, is required as a retrieval source.
The environmental map may cause a temporal and spatial gap between the retrieved information and the real world.
Therefore, the research topic in which the robots autonomously perceive the actual dynamic surroundings by using visual sensors mounted on the robot has been receiving a great deal of attention.

In particular, visible light sensors, also known as RGB cameras, are commonly used.
In the field of place categorization, Zhou \textit{et al.} proposed a dataset named Places~\cite{zhou2014learning} in 2014 and Places2~\cite{zhou2017places} in 2017 for benchmark evaluation.
The datasets are composed of color scenery images from hundreds of indoor and outdoor places and are preferable for training convolutional neural networks (CNNs) which are becoming the de facto standard framework for various image understanding tasks.
However, as sunlight decreases during nighttime and rainy weather, the appearance of outdoor scenes change dramatically and possibly cause the disappearance of discriminative visual attributes on the images.
This image diversity can easily lead to deterioration of performance and difficulties for learning networks.

One way to overcome image diversity is to fuse with other modalities.
Several researchers~\cite{Song:2015js, Silberman:2012kh, Mozos2012} have provided sequences of aligned colors and point clouds, i.e., RGB-D data, by using low-cost sensors such as Microsoft Kinect and ASUS Xtion.
However, the errors and defects that tend to occur in data are due to sensor characteristics, and previous research has been limited to indoor applications.
Recent research is trending towards outdoor scene understanding and includes increasing use of 3D data from light detection and ranging sensors (LiDARs) to improve depth and directional range while ensuring robust to illumination.
Nevertheless, most studies focus on not place categorization but on localization and mapping problems motivated by autonomous driving, and the 3D datasets are proposed for these problems~\cite{blanco2014ijrr,fordcampus}.
Only the KITTI~\cite{geiger2012we} dataset includes four categories in scans annotated manually: city, residential, road, and campus.
This dataset is used as a benchmark for other purposes such as optical flow estimation, visual odometry, 3D object detection, and 3D tracking.

In this research, we first construct two types of multi-modal 3D datasets, named ``Multi-modal Panoramic 3D Outdoor (MPO)'', for outdoor place categorization.
Our datasets are recorded by multi-modal sensors, provided as multi-resolution point clouds, and composed of six place categories.
In addition, we propose new CNN architectures extended for the multi-modal 3D data obtained from a single LiDAR.
First, we convert the point clouds into panoramic image representation that is more compact than raw 3D structure.
Moreover, we use the reflection intensity (reflectance) maps obtained simultaneously as by-products of time-of-flight distance measurement, which are fully aligned to the range (depth) maps.

\begin{figure}[t]
    \centering
    \subfigure[Panoramic color image of the urban scene]
    {
        \includegraphics[width=0.8\hsize]{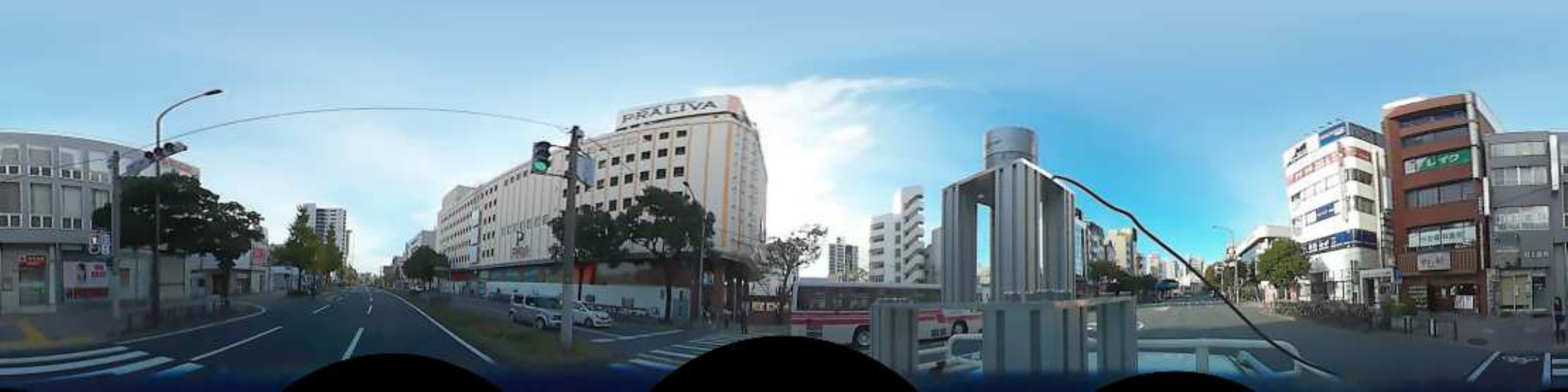}
    }
    \subfigure[3D point clouds obtained from the LiDAR]
    {
        \includegraphics[width=0.8\hsize]{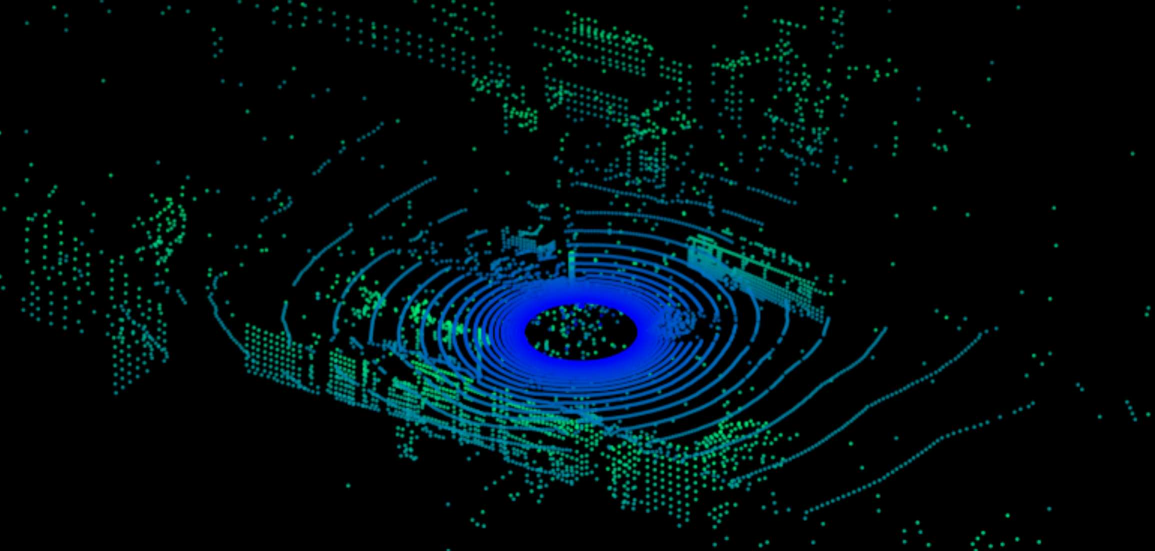}
    }
    \subfigure[Corresponding panoramic depth image]
    {
        \includegraphics[width=0.8\hsize]{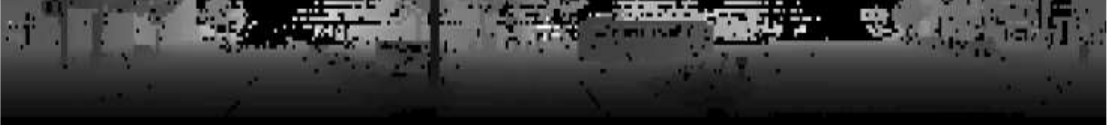}
    }
    \subfigure[Corresponding panoramic reflectance image]
    {
        \includegraphics[width=0.8\hsize]{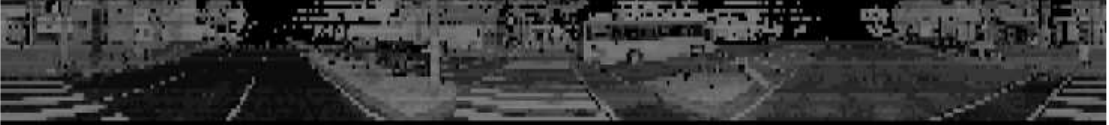}
    }
    \caption{Samples from our dataset. The panoramic images represent the surrounding geometric and photometric features compactly.}
    \label{fig:sample}
\end{figure}

The rest of this paper is organized as follows. In Section~\ref{sec:related_work} we describe related work in terms of data to process, approach, and problems.
In Section ~\ref{sec:dataset} we introduce the MPO datasets and describe the collection procedure.
In Section ~\ref{sec:models} we introduce our proposed uni-modal models and multi-modal models.
Finally, we show the experimental results in Section~\ref{sec:experiment} and state our conclusion in Section~\ref{sec:conclusion}.
\section{Related Work}
\label{sec:related_work}

Outdoor place recognition has been studied in computer vision and robotics, and it is now highly motivated because its application to outdoor autonomous robots and vehicles.

The place recognition tasks have traditionally been cast as instance-level recognition to identify the location where an image is captured or category-level recognition to answer the general attribute of a place.
As an example of instance-level recognition, Torii \textit{et al.}~\cite{torii201524} densely extracted SIFT~\cite{lowe2004distinctive} features from the query image and retrieves the most similar image from a view-synthesized database. Conversely, our aim is the category-level recognition, that targets unknown places and predicts their semantic categories.
Related to this viewpoint, Xiao \textit{et al.}~\cite{xiao2010sun} applied several local and global feature techniques for place images. Fazl-Ersi \textit{et al.}~\cite{fazl2012ijrr} applied histograms of oriented uniform patterns of images in outdoor environments.
Alternatively, depth information can be used to categorize places. The work in~\cite{Mozos2005} extracted geometrical features from 2D laser scans to categorize indoor places. In addition, 3D depth information has been used to categorize indoor places by using RGB-D sensors~\cite{moral2013icra} and 3D laser scans~\cite{Mozos2013}.

A common problem in traditional techniques was how to extract efficient visual features. Recently, CNNs have gained a great deal of attention as a powerful method to automatically extract the features in visual recognition tasks.
For instance-level recognition using color images, Arandjelovi\'{c} \textit{et al.}~\cite{arandjelovic2016netvlad} proposed a CNN architecture to recognize place instances by treating the problem as an image retrieval task. Gomez-Ojeda \textit{et al.}~\cite{gomez2015training} trained a CNN based on the triplet loss calculated from three instances, for the purpose of recognizing revisited places under significant appearance changes. S\"underhauf \textit{et al.}~\cite{sunderhauf2015performance} investigated the performance of a CNN as an image descriptor and its robustness to appearance and viewpoint changes. For category-level recognition, Zhou \textit{et al.}~\cite{zhou2014learning} used a CNN for scene recognition in the Places 205 dataset. Ur\u{s}i\u{c} \textit{et al.}~\cite{UrsicICRA17} proposed a part-based model of household space categories based on CNNs.

While several methods have used depth images to recognize object categories~\cite{maturana2015voxnet}, to estimate object shapes~\cite{wu20153d}, or to detect vehicles for autonomous driving~\cite{li2016vehicle}, not many works have focused on depth images to solve the place categorization task by using CNNs.
Sizikova \textit{et al.}~\cite{sizikova2016enhancing} used generated synthetic 3D data to train a CNN for indoor place recognition; however, they presented this task as an image matching problem at the instance level.
For works related to ours, Goeddel \textit{et al.}~\cite{GoeddelIROS2016} transformed 2D laser reading into an image and train CNNs to classify household places such as a room, a corridor, or a doorway. Song \textit{et al.}~\cite{Song:2015js} proposed the SUN RGB-D indoor scene database and performed scene categorization by concatenating color-based and depth-based CNN features. Still, these works have been limited to indoor applications.

In this paper, we integrate CNNs and large-scale LiDAR data which includes a range and reflectance modalities and aim to predict generic categories in outdoor places by automatically learning geometric and photometric features. Because the multi-modal data from a LiDAR are more robust to changes in illumination, they have a big advantage for autonomous vehicles navigating outdoors.
\section{MPO Datasets}
\label{sec:dataset}
We present Multi-modal Panoramic 3D Outdoor (MPO), a dataset for outdoor place categorization.
The dataset consists of two subsets with different scenarios, \textit{sparse} and \textit{dense}.
For both sets, we acquired data in the following six types of places: coast, forest, indoor parking, outdoor parking, residential area, and urban area.
The datasets are available on the web~\footnote{Dense Multi-modal Panoramic 3D Outdoor Dataset for Place Categorization, \url{http://robotics.ait.kyushu-u.ac.jp/~kurazume/research-e.php?content=db#d07}, 2016}~\footnote{Sparse Multi-modal Panoramic 3D Outdoor Dataset for Place Categorization, \url{http://robotics.ait.kyushu-u.ac.jp/~kurazume/research-e.php?content=db#d08}, 2016}.
In this paper we only use the sparse set for training and evaluation of CNNs, because of that the amount of the dense set is small to train from scratch.

\subsection{Sensors and Data}

\subsubsection{Point Clouds}
The omnidirectional laser scanner, LiDAR, is widely used as a sensor for collecting large-scale point clouds representing geometric information of outdoor environments.
A LiDAR obtains 360-degree range data by rotating a multi-line scanner head around the vertical axis.
The range, reflectance, and polar/azimuth angle in spherical coordinates are given for each measured point.
We used two kinds of LiDARs with different resolutions.
We provide details about the sensors in Section~\ref{sec:data_acq}.

\begin{figure}[htb]
    \centering
    \subfigure[Velodyne HDL-32e]{
        \includegraphics[width=0.2\hsize]{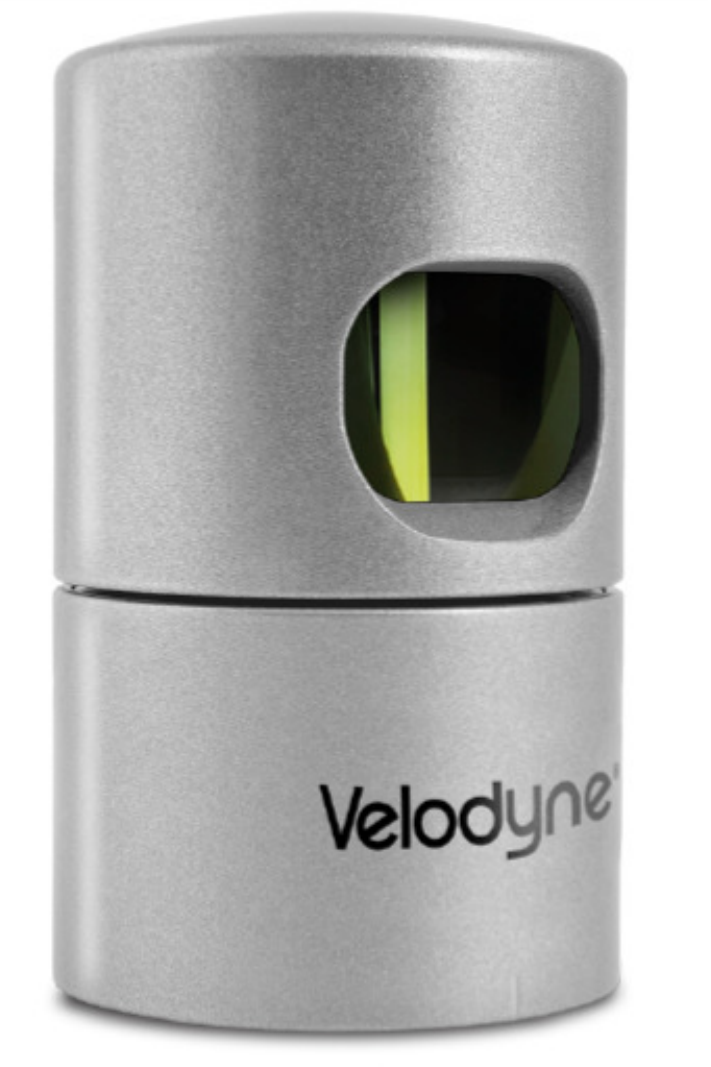}
    }
    \subfigure[GARMIN GPS 18x LVC]{
        \includegraphics[width=0.28\hsize]{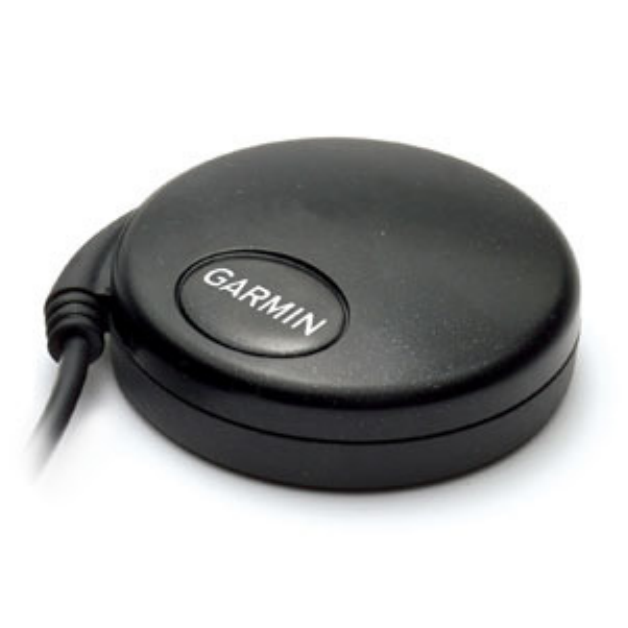}
    }
    \subfigure[Kodak PIXPRO SP360]{
        \includegraphics[width=0.2\hsize]{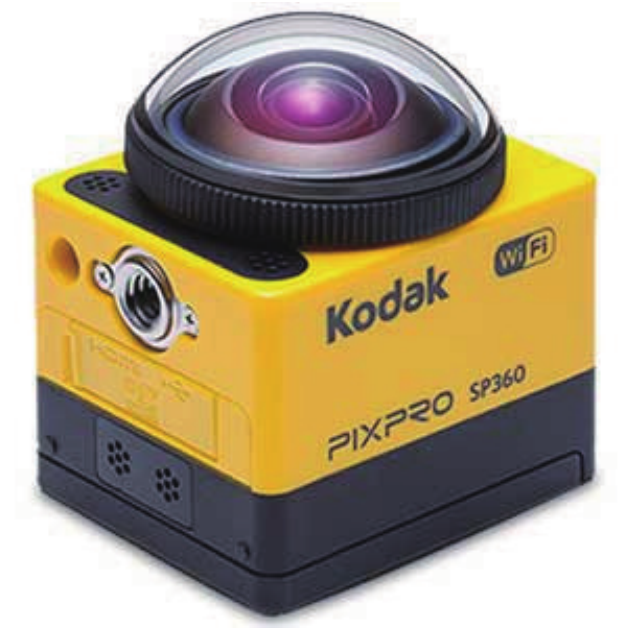}
    }
    \caption{Omnidirectional laser scanner, GPS, and 360 deg. camera}
    \label{fig:sensors}
\end{figure}

\subsubsection{Geolocation}
To confirm the locations of laser scanning, we stored the GPS data with the point clouds simultaneously.
The GPS sensor we used is Garmin GPS 18x LVC (Fig.~\ref{fig:sensors}(b)).
The specifications are shown in Table~\ref{table:gps}.
We obtained location data consisting of latitude/longitude location, date/time stamp, and speed over ground in the National Marine Electronics Association (NMEA) \texttt{\$GPRMC} data format.

\subsubsection{Omnidirectional Images}
The FARO sensor in the dense set can capture a color image with a synchronized depth image.
For the sparse set, we acquired omnidirectional color images by an additional camera, Kodak PIXPRO SP360 (Fig.~\ref{fig:sensors}(c)).
Each color image covers 360 degrees and has a resolution of 16.36 megapixels.

\subsection{Data Acquisition}
\label{sec:data_acq}

\subsubsection{Sparse MPO}
\begin{figure}[htb]
    \centering
    \subfigure[Sensor setup]{
        \includegraphics[height=0.28\hsize]{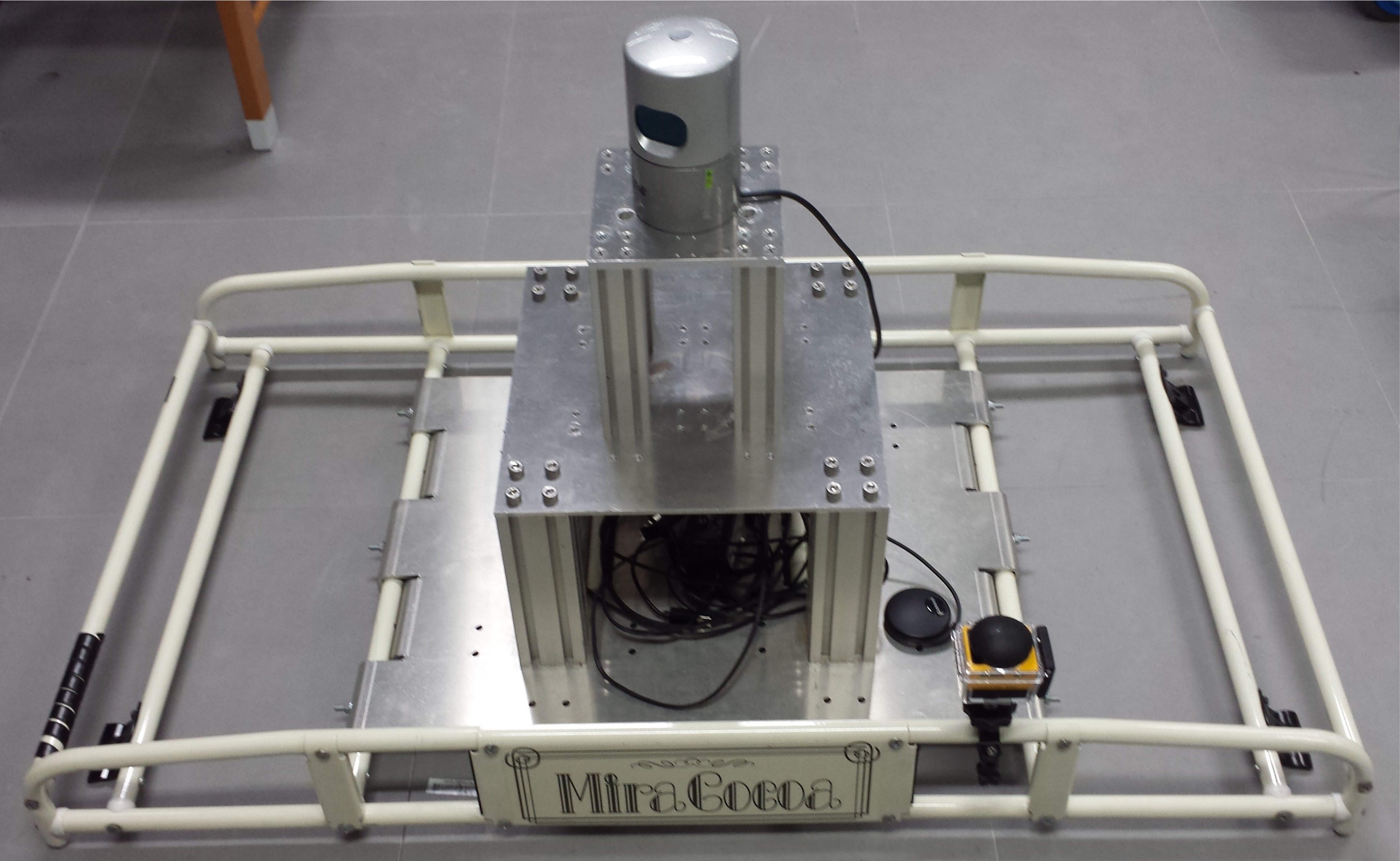}
    }
    \subfigure[Sensors mounted on a car]{
        \includegraphics[height=0.28\hsize]{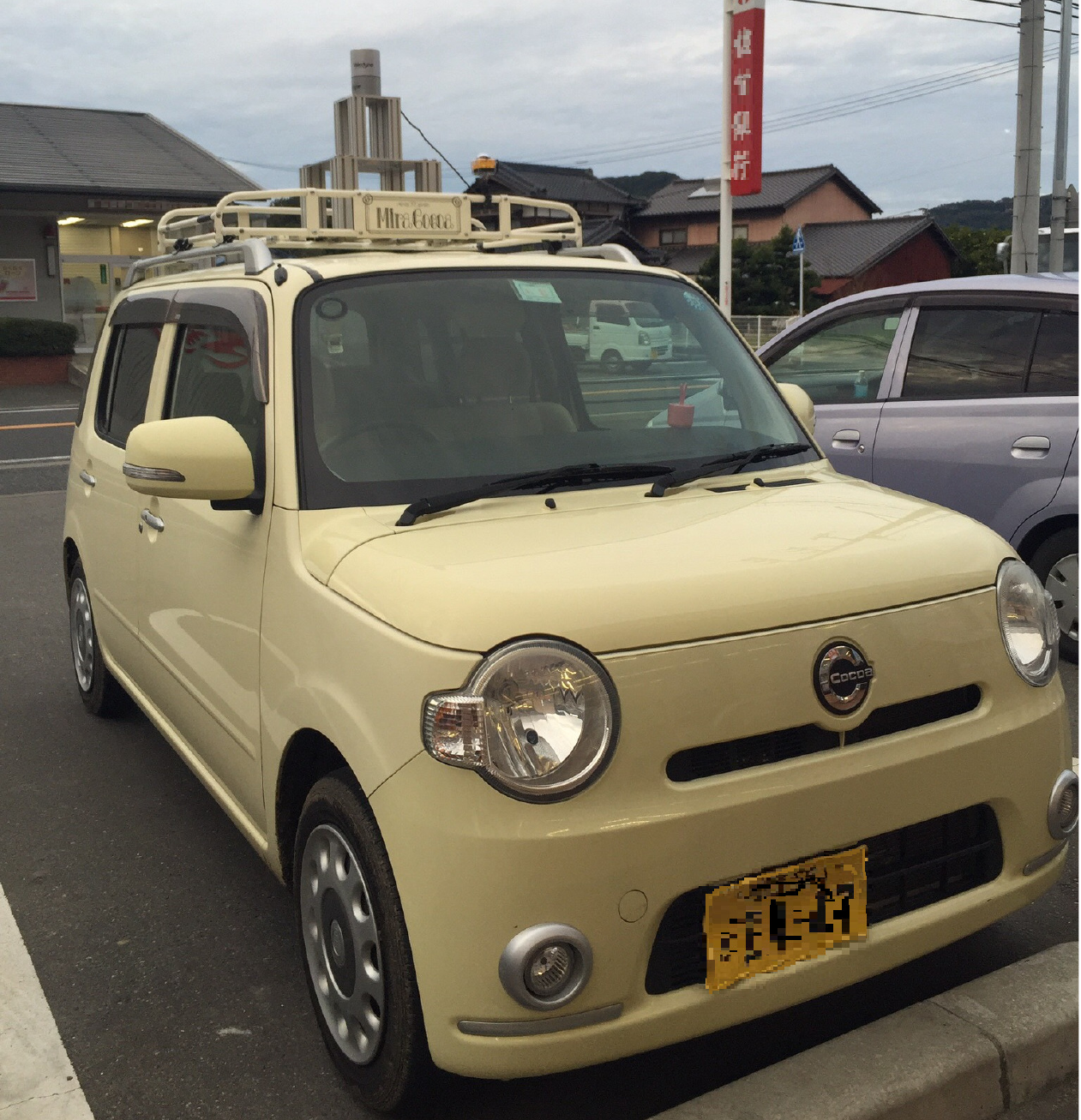}
    }
    \caption{Data acquisition setup}
    \label{fig:car}
\end{figure}

\begin{figure}[htb]
    \centering
    \includegraphics[width=0.32\textwidth]{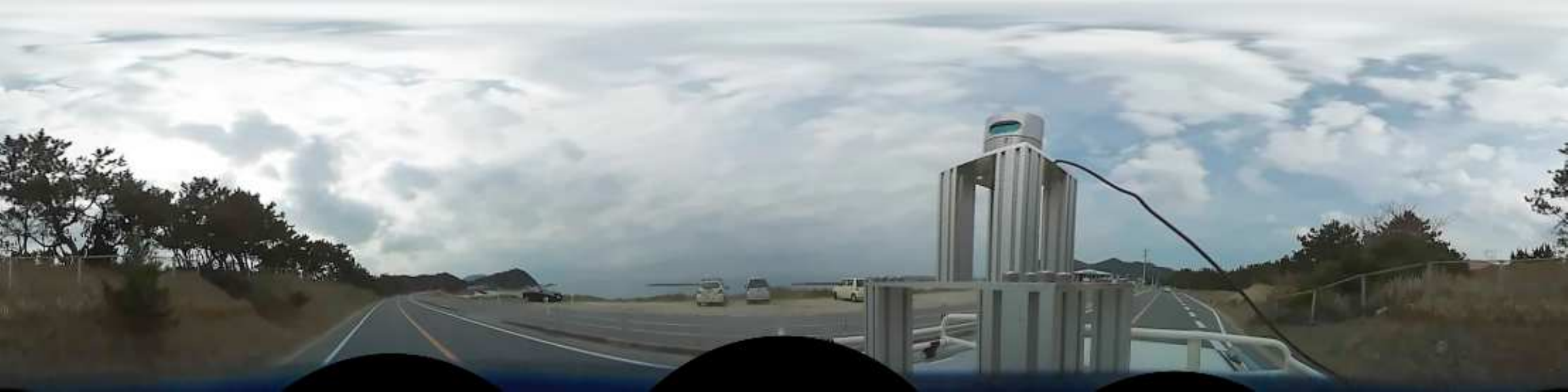} \hfill
    \includegraphics[width=0.32\textwidth]{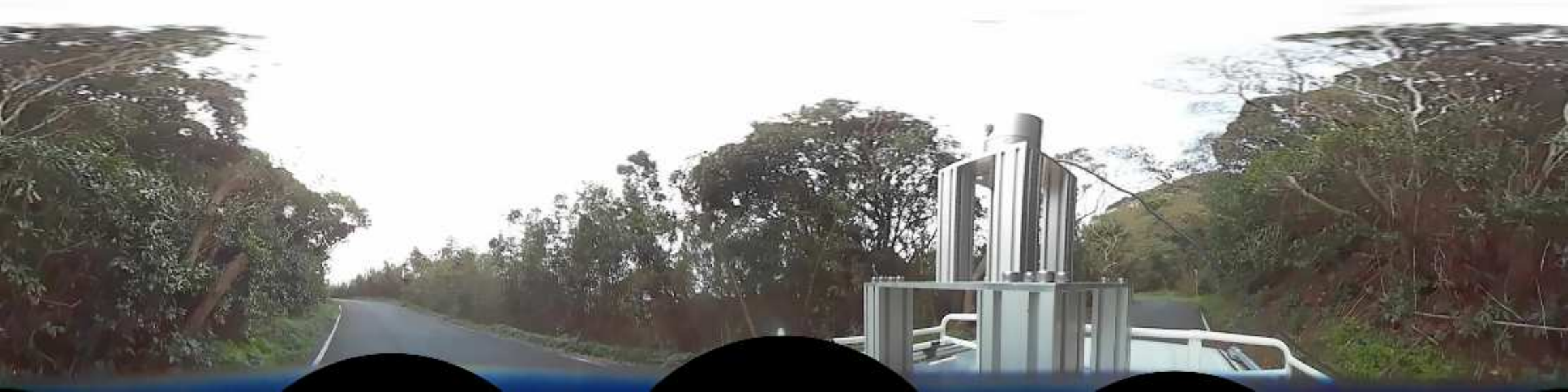} \hfill
    \includegraphics[width=0.32\textwidth]{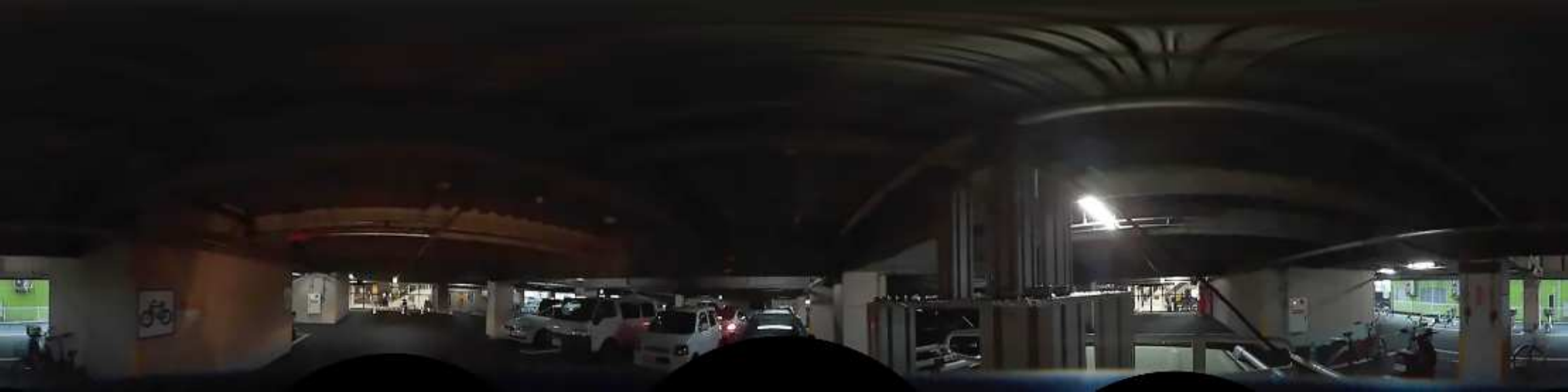}
    \\
    \includegraphics[width=0.32\textwidth]{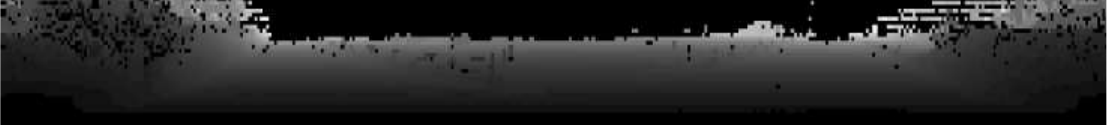} \hfill
    \includegraphics[width=0.32\textwidth]{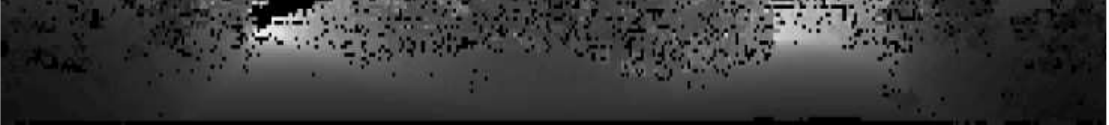} \hfill
    \includegraphics[width=0.32\textwidth]{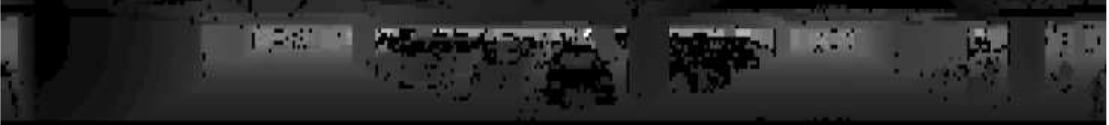}
    \\
    \includegraphics[width=0.32\textwidth]{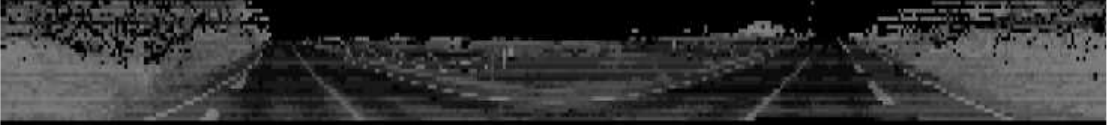} \hfill
    \includegraphics[width=0.32\textwidth]{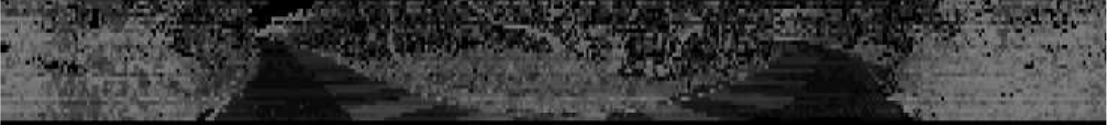} \hfill
    \includegraphics[width=0.32\textwidth]{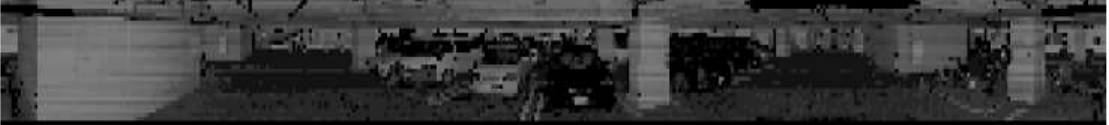}
    \\
    \begin{minipage}[c]{0.32\textwidth}
        \centering
        {\footnotesize Coast}
    \end{minipage}
    \hfill
    \begin{minipage}[c]{0.32\textwidth}
        \centering
        {\footnotesize Forest}
    \end{minipage}
    \hfill
    \begin{minipage}[c]{0.32\textwidth}
        \centering
        {\footnotesize ParkingIn}
    \end{minipage}
    \vspace{3mm}
    \\
    \includegraphics[width=0.32\textwidth]{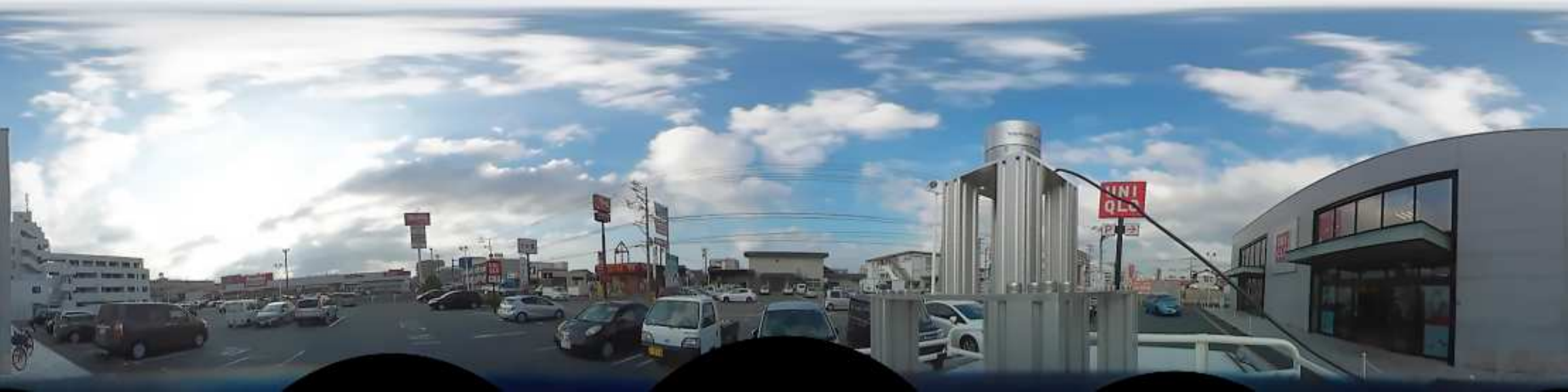} \hfill
    \includegraphics[width=0.32\textwidth]{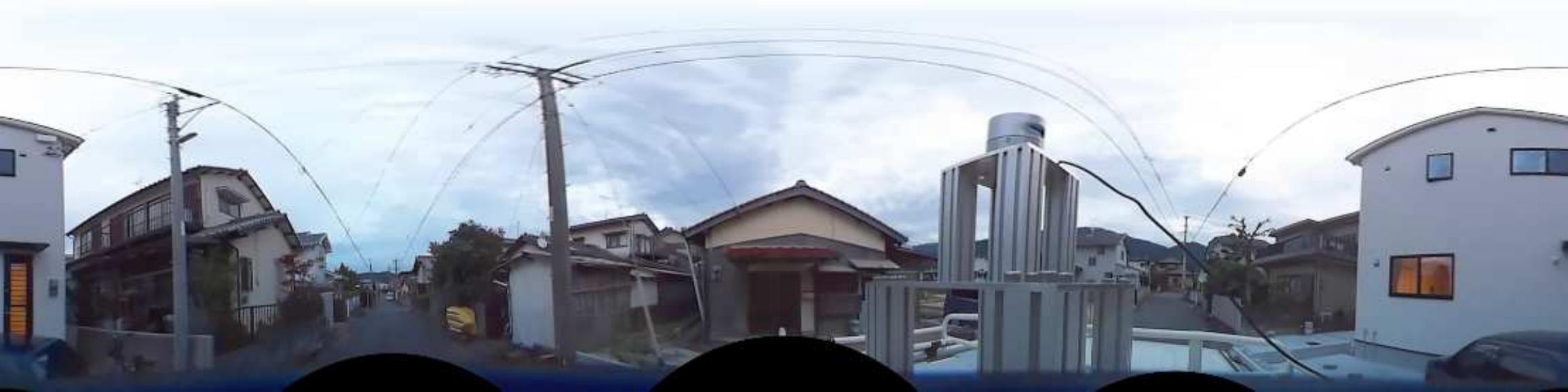} \hfill
    \includegraphics[width=0.32\textwidth]{figure/SparseMPO/sample/Urban_RGB.pdf}
    \\
    \includegraphics[width=0.32\textwidth]{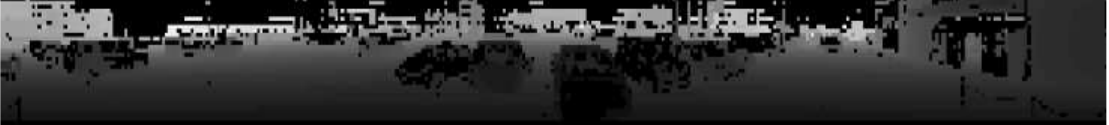} \hfill
    \includegraphics[width=0.32\textwidth]{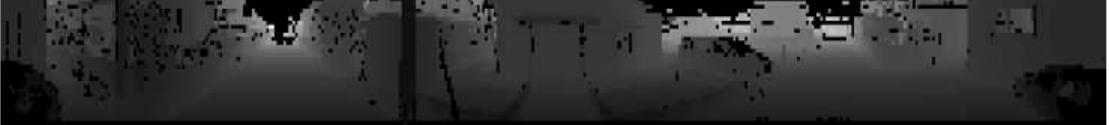} \hfill
    \includegraphics[width=0.32\textwidth]{figure/SparseMPO/sample/Urban_Dep.pdf}
    \\
    \includegraphics[width=0.32\textwidth]{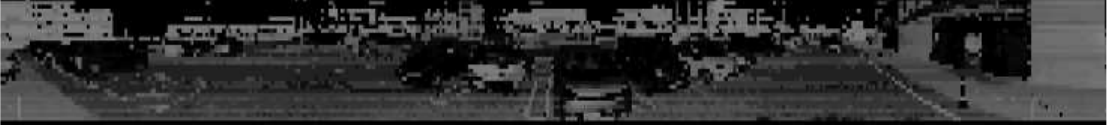} \hfill
    \includegraphics[width=0.32\textwidth]{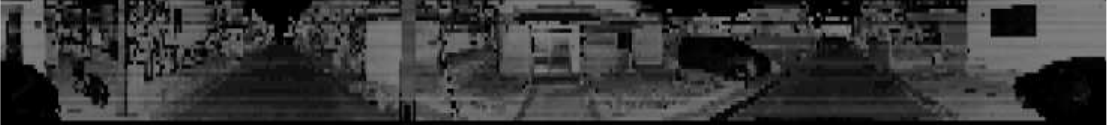} \hfill
    \includegraphics[width=0.32\textwidth]{figure/SparseMPO/sample/Urban_Ref.pdf}
    \\
    \begin{minipage}[c]{0.32\textwidth}
        \centering
        {\footnotesize ParkingOut}
    \end{minipage}
    \hfill
    \begin{minipage}[c]{0.32\textwidth}
        \centering
        {\footnotesize Residential}
    \end{minipage}
    \hfill
    \begin{minipage}[c]{0.32\textwidth}
        \centering
        {\footnotesize Urban}
    \end{minipage}
    \caption{Examples of an RGB (top), depth (middle), and reflectance image (bottom) for each place category.}
    \label{fig:sparsempo_sample}
\end{figure}

\begin{figure}[htb]
    \centering
    \includegraphics[width=0.8\hsize]{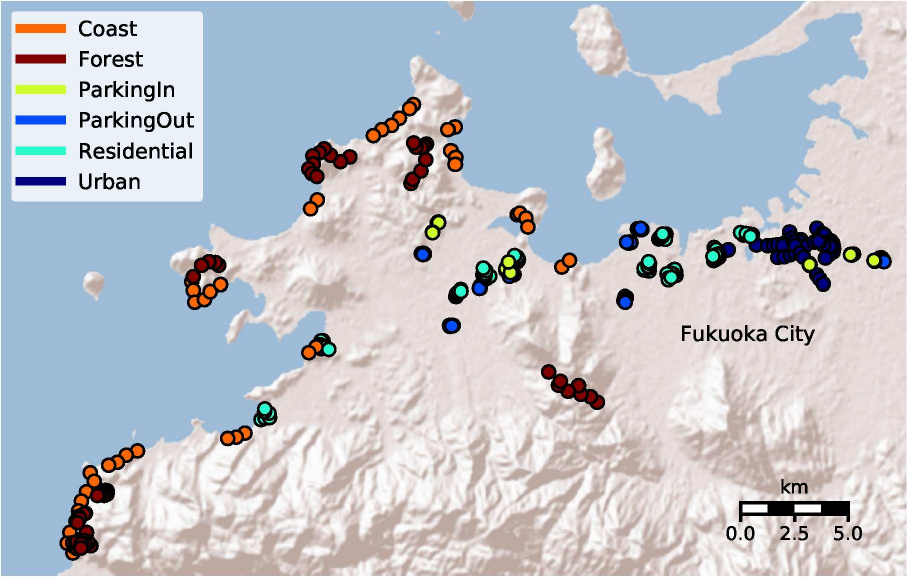}
    \caption{Scanning locations for Sparse Multi-modal Panoramic 3D Outdoor Dataset}
    \label{fig:gmap}
\end{figure}

We used the Velodyne HDL-32e (Fig.~\ref{fig:sensors}(a)) as the laser scanner. The specifications of the Velodyne HDL-32e are shown in Table~\ref{table:velodyne}.
The laser scanner, the GPS sensor, and the 360-deg camera are connected with the robot operating system (ROS), and so color images, range data, and GPS data can be synchronized by using timestamps in the ROS message.
We mounted the set of sensors on the top of a car (Fig.~\ref{fig:car}).
The velocity of the car was about 30 km/h to 50 km/h and all data were recorded at 2 Hz.
For each place category we defined, we chose ten areas in Fukuoka city, Japan, and scanned the environments while driving the car.
Thus each data was annotated with one out of six types of place categories on a per-area basis.
Some examples of the depth and reflectance images for each place category are shown in Fig.~\ref{fig:sparsempo_sample}.
The Velodyne HDL-32e scans 2166 points on each of 32 vertical channels, thus the resolution of the generated images is $2166\times32$.
Table~\ref{table:sparsempo} shows the total number of images is 34,200 and the total amount of data is 59.23 GB.
In addition, the locations given by the GPS sensor are shown on the map in Fig.~\ref{fig:gmap}.

\begin{table}[htb]
    \tbl{Distribution on Sparse-MPO}
    {
        \centering
        \begin{tabular}{llllllllllll}
            \toprule
            Category                                             & \multicolumn{10}{l}{Number of scans at each location} & Total                                                                                                                                        \\ \cmidrule{2-11}
                        & {\tiny Set1} & {\tiny Set2} & {\tiny Set3} & {\tiny Set4} & {\tiny Set5} & {\tiny Set6} & {\tiny Set7} & {\tiny Set8} & {\tiny Set9} & {\tiny Set10} &      \\
            \midrule
            Coast       & 511          & 254          & 571          & 221          & 314          & 376          & 872          & 506          & 386          & 287           & 4298 \\
            \midrule
            Forest      & 440          & 824          & 980          & 707          & 730          & 720          & 439          & 311          & 797          & 531           & 6479 \\
            \midrule
            ParkingIn   & 520          & 357          & 274          & 873          & 583          & 343          & 466          & 592          & 344          & 428           & 4780 \\
            \midrule
            ParkingOut  & 874          & 579          & 388          & 370          & 477          & 536          & 581          & 563          & 460          & 617           & 5445 \\
            \midrule
            Residential & 674          & 787          & 667          & 724          & 563          & 973          & 717          & 720          & 977          & 662           & 7464 \\
            \midrule
            Urban       & 490          & 572          & 587          & 487          & 410          & 566          & 712          & 565          & 606          & 739           & 5734 \\
            \midrule
            \multicolumn{11}{r}{Total number of panoramic scans} & 34200                                                                                                                                                                                                \\
            \bottomrule
        \end{tabular}
    }
    \label{table:sparsempo}
\end{table}

\subsubsection{Dense MPO}
\begin{figure}[htb]
    \centering
    \includegraphics[width=0.8\hsize]{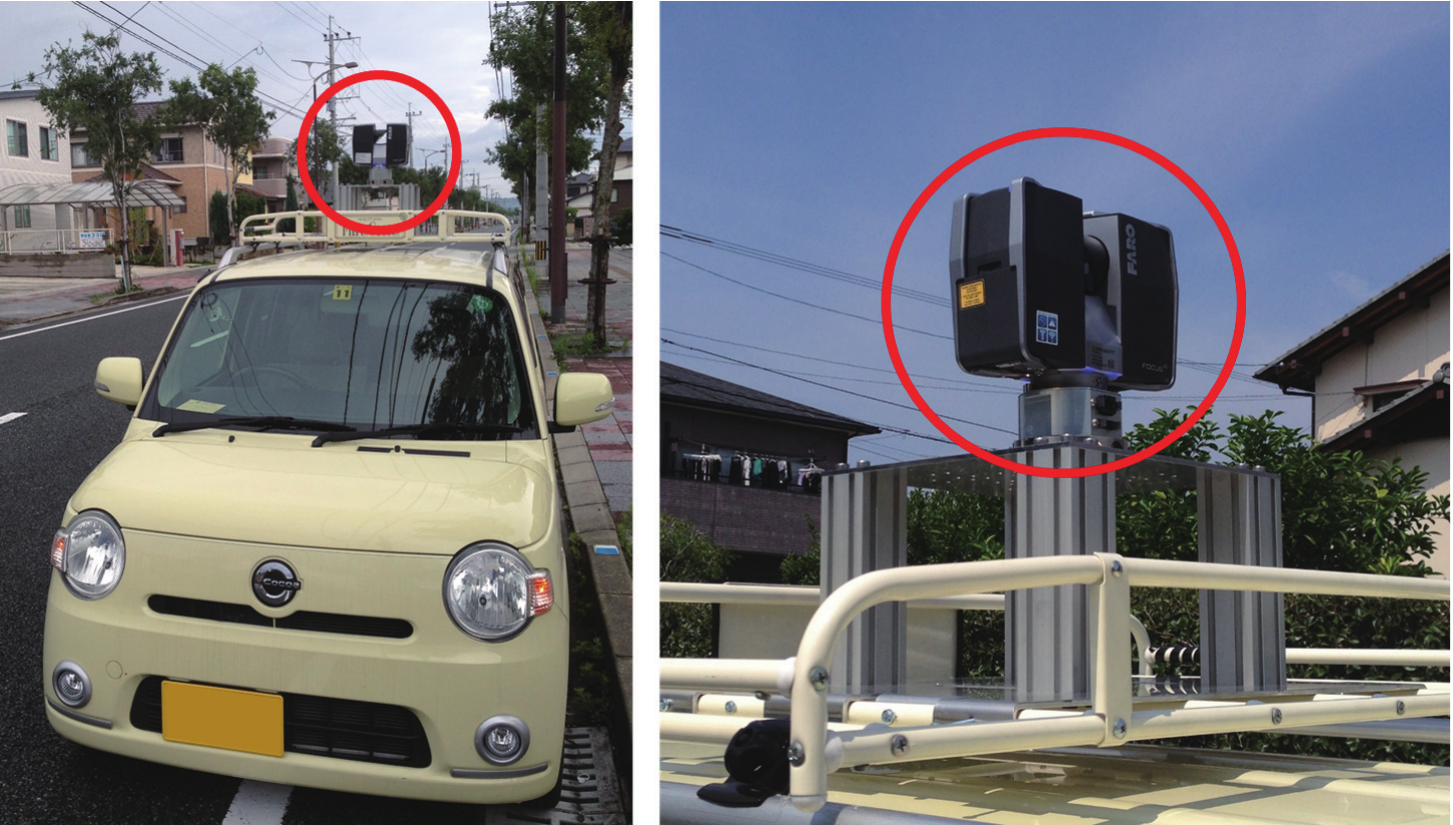}
    \caption{FARO Focus 3D S120}
    \label{fig:FARO}
\end{figure}

\begin{figure}[htb]
    \centering
    \includegraphics[width=0.32\textwidth]{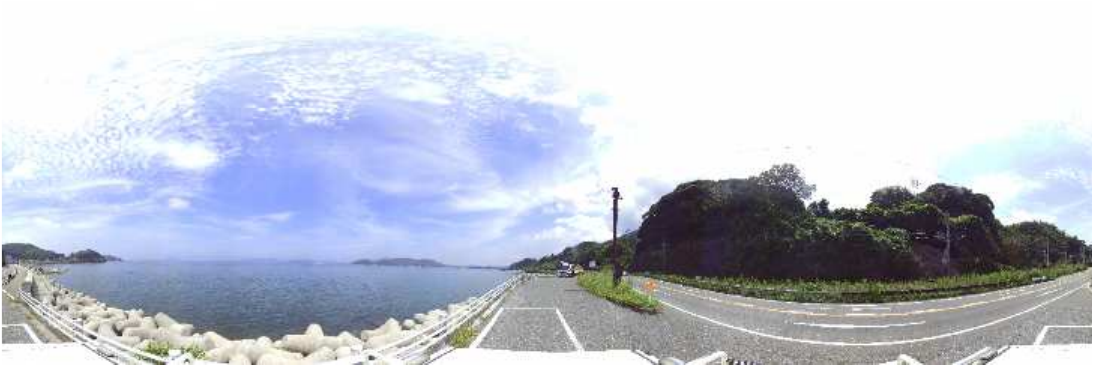} \hfill
    \includegraphics[width=0.32\textwidth]{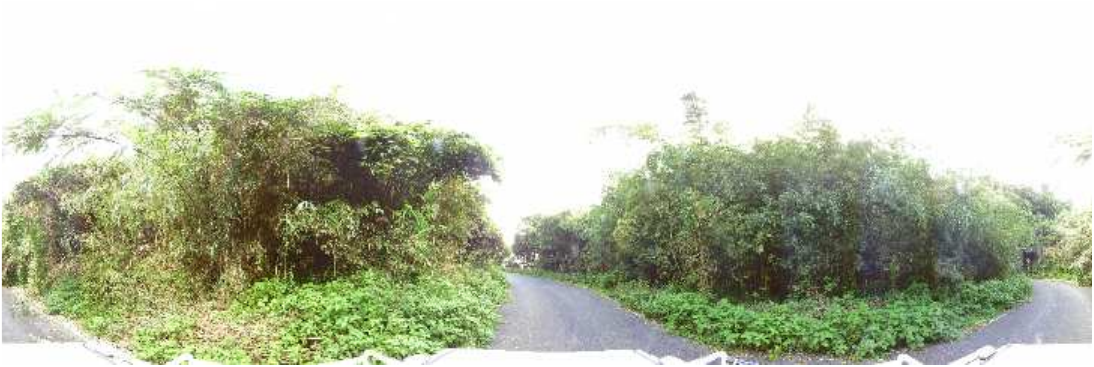} \hfill
    \includegraphics[width=0.32\textwidth]{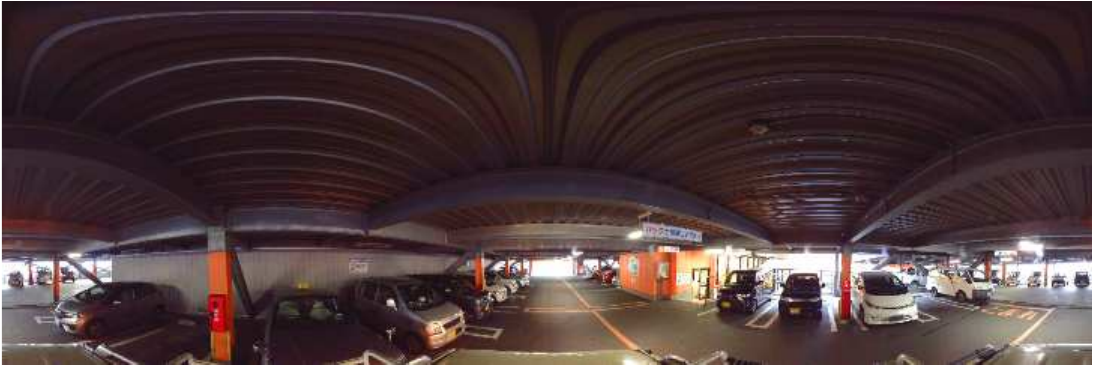}
    \\
    \includegraphics[width=0.32\textwidth]{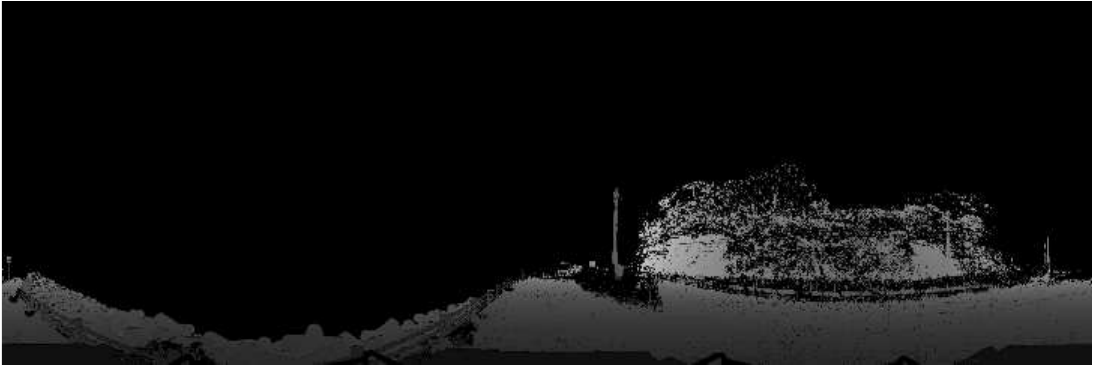} \hfill
    \includegraphics[width=0.32\textwidth]{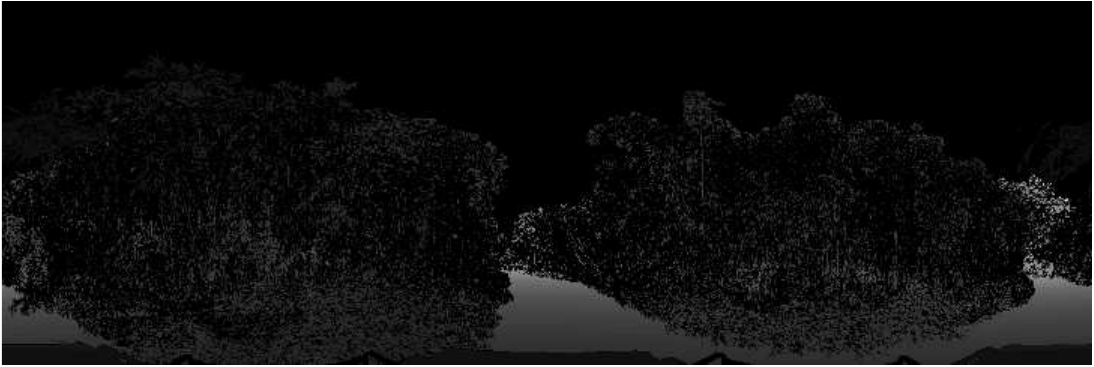} \hfill
    \includegraphics[width=0.32\textwidth]{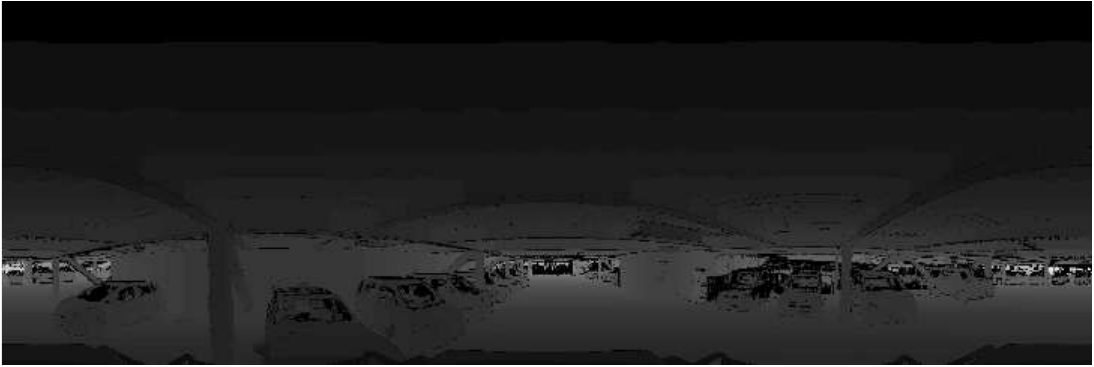}
    \\
    \includegraphics[width=0.32\textwidth]{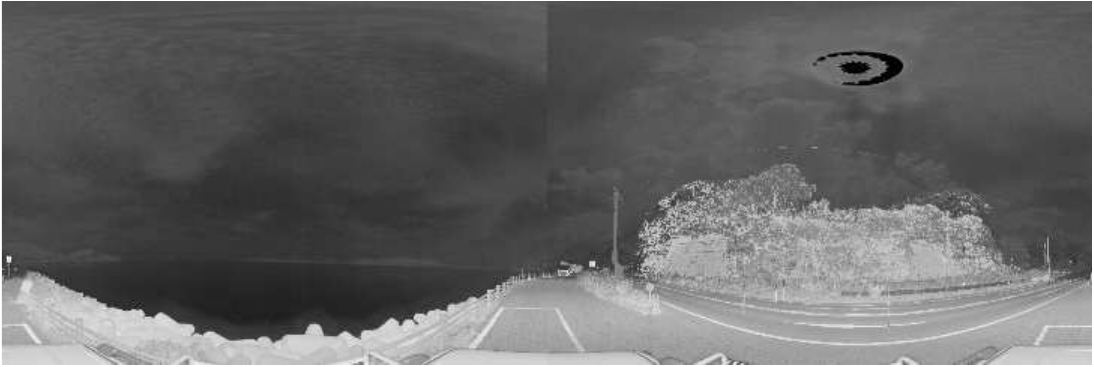} \hfill
    \includegraphics[width=0.32\textwidth]{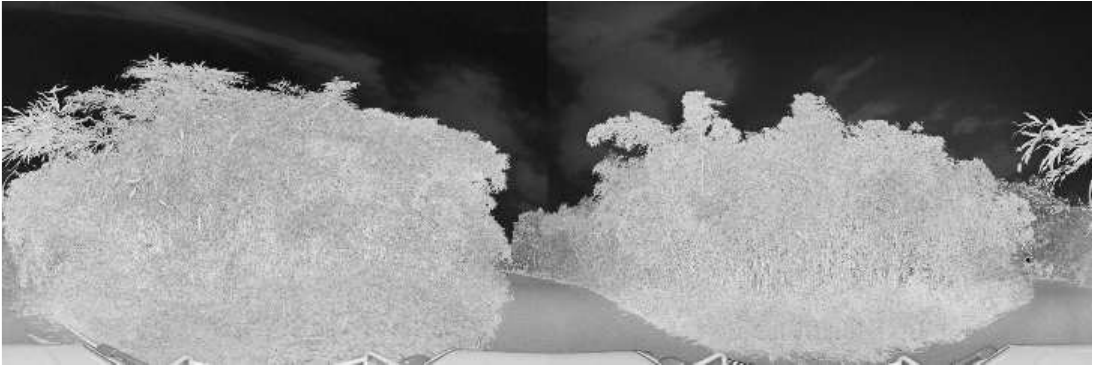} \hfill
    \includegraphics[width=0.32\textwidth]{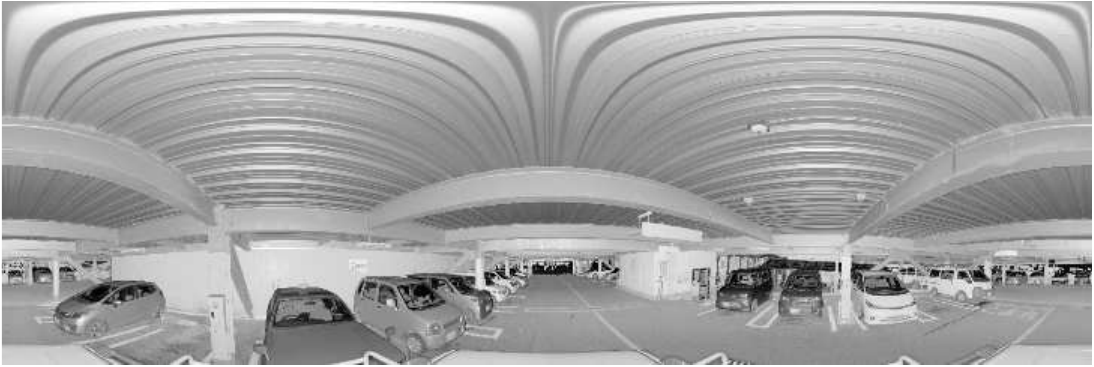}
    \\
    \begin{minipage}[c]{0.32\textwidth}
        \centering
        {\footnotesize Coast}
    \end{minipage}
    \hfill
    \begin{minipage}[c]{0.32\textwidth}
        \centering
        {\footnotesize Forest}
    \end{minipage}
    \hfill
    \begin{minipage}[c]{0.32\textwidth}
        \centering
        {\footnotesize ParkingIn}
    \end{minipage}
    \vspace{3mm}
    \\
    \includegraphics[width=0.32\textwidth]{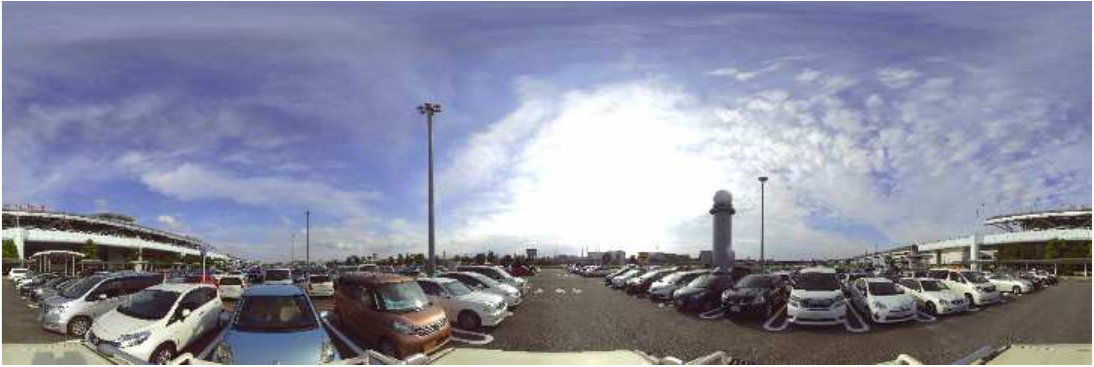} \hfill
    \includegraphics[width=0.32\textwidth]{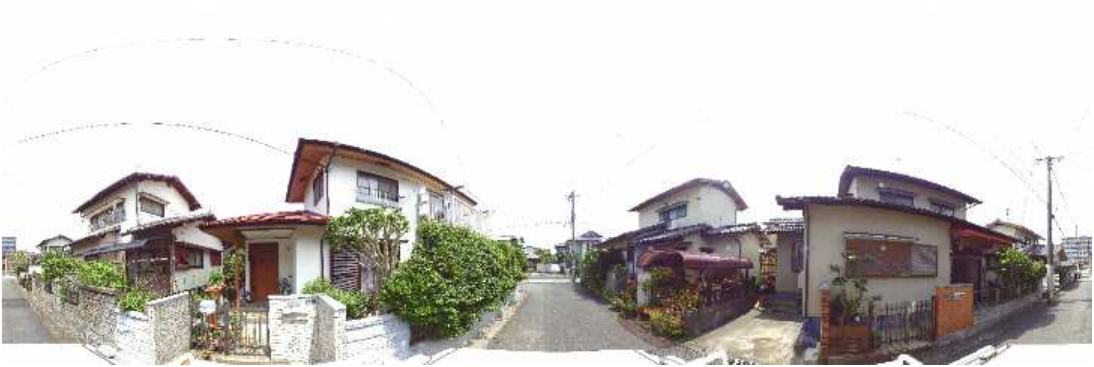} \hfill
    \includegraphics[width=0.32\textwidth]{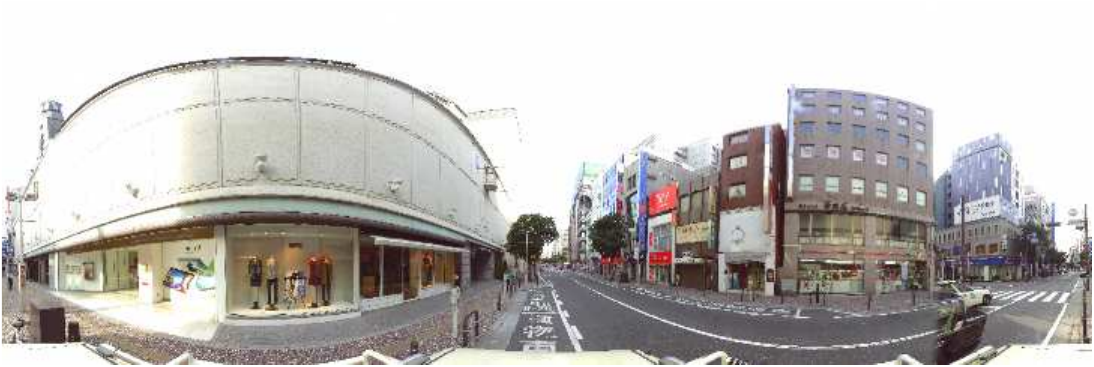}
    \\
    \includegraphics[width=0.32\textwidth]{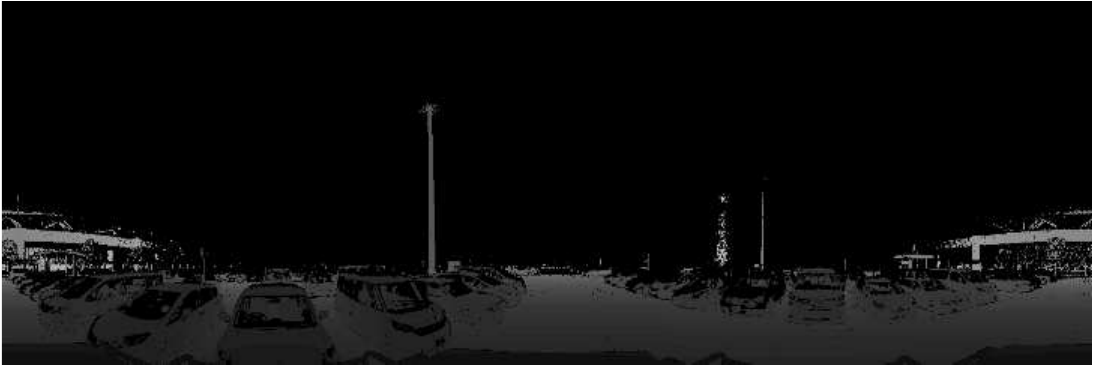} \hfill
    \includegraphics[width=0.32\textwidth]{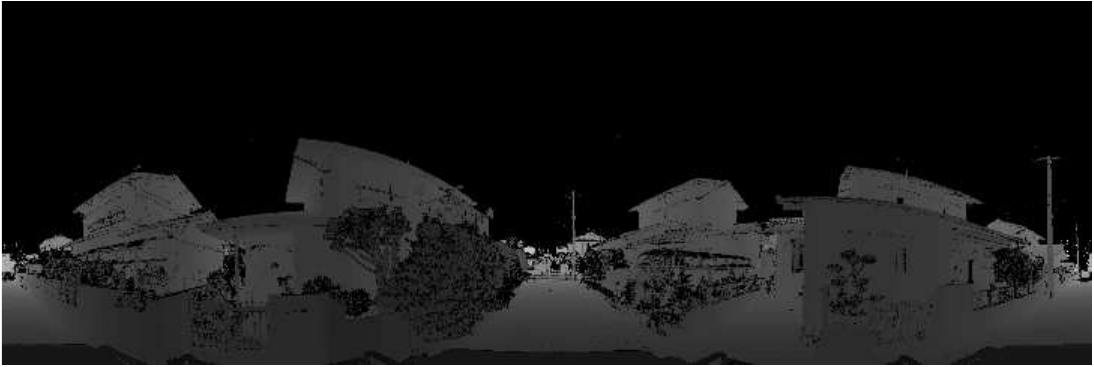} \hfill
    \includegraphics[width=0.32\textwidth]{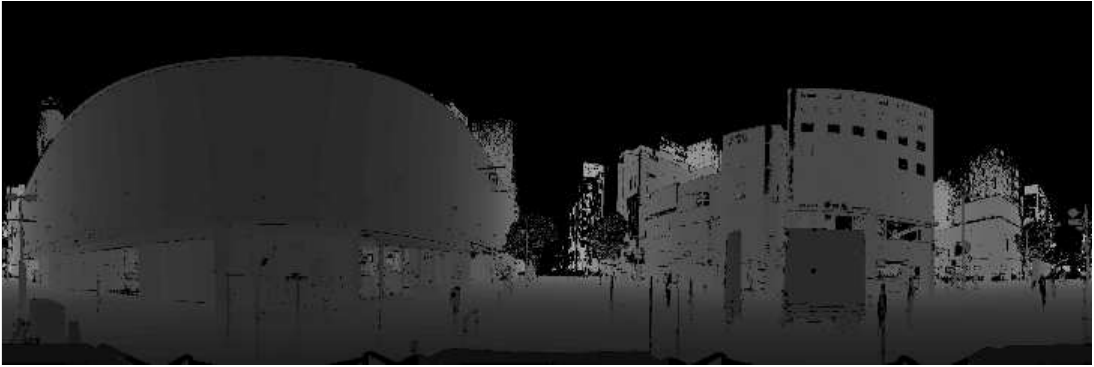}
    \\
    \includegraphics[width=0.32\textwidth]{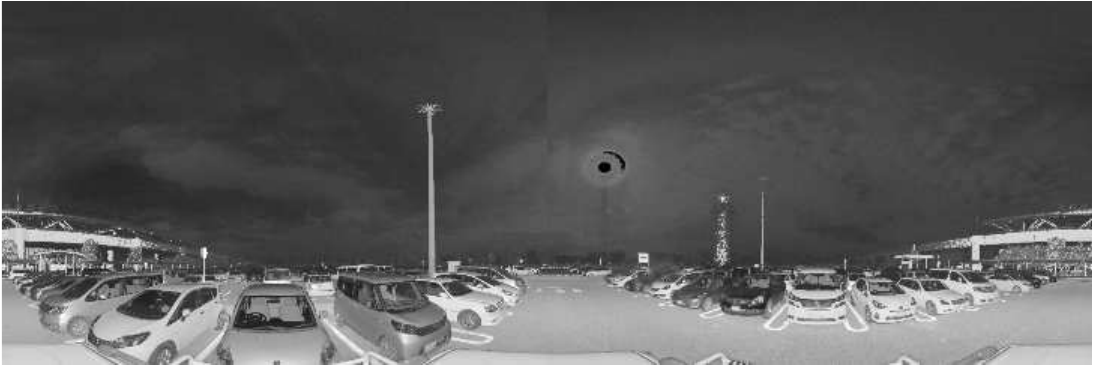} \hfill
    \includegraphics[width=0.32\textwidth]{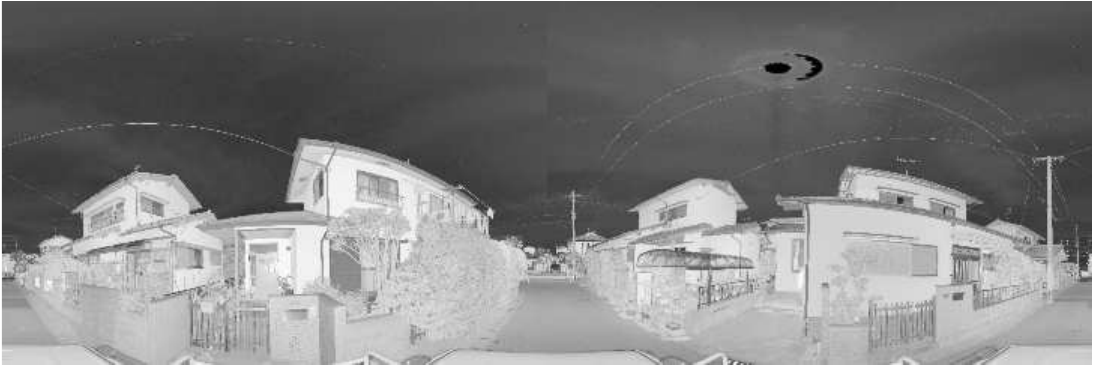} \hfill
    \includegraphics[width=0.32\textwidth]{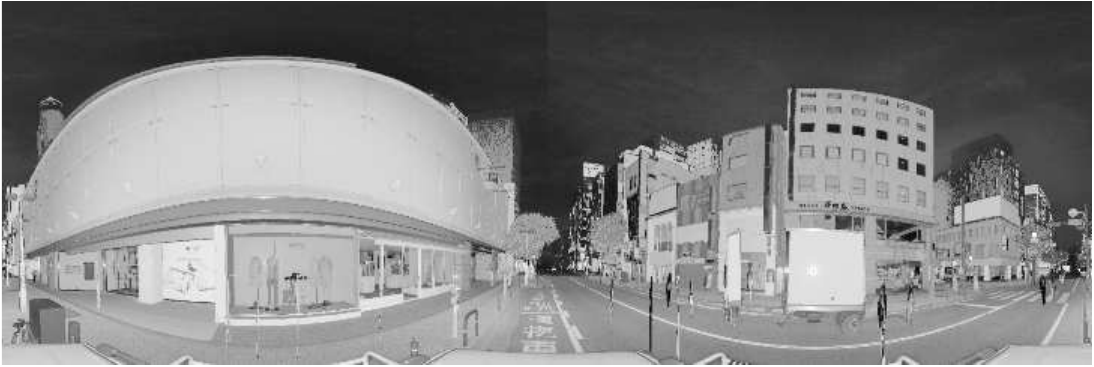}
    \\
    \begin{minipage}[c]{0.32\textwidth}
        \centering
        {\footnotesize ParkingOut}
    \end{minipage}
    \hfill
    \begin{minipage}[c]{0.32\textwidth}
        \centering
        {\footnotesize Residential}
    \end{minipage}
    \hfill
    \begin{minipage}[c]{0.32\textwidth}
        \centering
        {\footnotesize Urban}
    \end{minipage}
    \caption{Examples of an RGB (top), depth (middle), and reflectance image (bottom) for each place category.}
    \label{fig:densempo_sample}
\end{figure}

\begin{figure}[t]
    \centering
    \includegraphics[width=0.8\hsize]{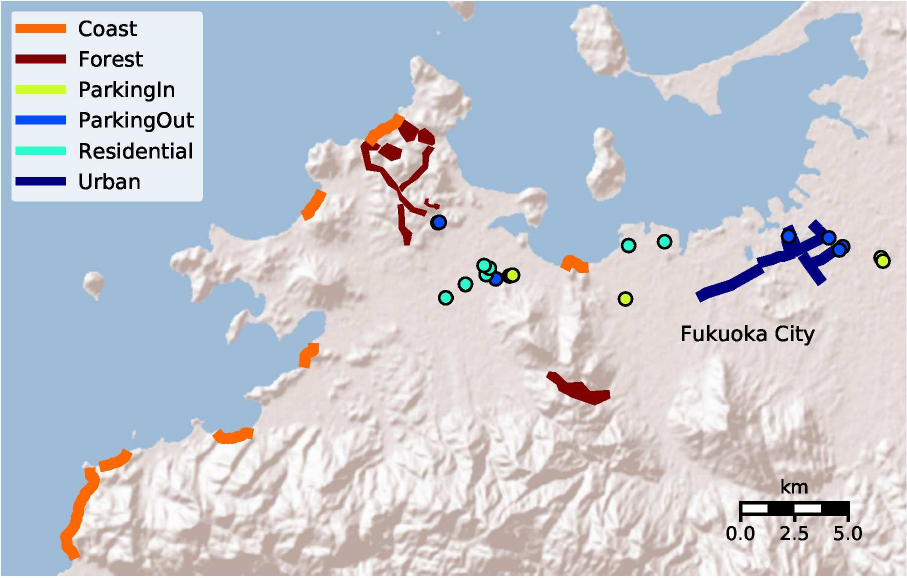}
    \caption{Scanning locations for Dense Multi-modal Panoramic 3D Outdoor Dataset}
    \label{fig:map_FARO}
\end{figure}

A high-resolution omnidirectional laser scanner, FARO Focus 3D S120 (Fig.~\ref{fig:FARO}) was used for constructing the dense set. The specifications of FARO Focus 3D S120 are shown in Table~\ref{table:FARO}.
We put the Focus 3D S120 on a car as shown in Fig.~\ref{fig:FARO}, moved and stopped in the seven areas shown in Fig.~\ref{fig:map_FARO}, and scanned the surroundings.
The measured data are panoramic depth images, panoramic reflectance images, and panoramic color images.    Like the sparse set, each data was annotated with one out of six types of place categories on a per-area basis.
The resolution of each image is $5140\times1757$.
Table~\ref{table:densempo} shows the total number of images in the dataset is 650 and the size of data is about 242 GB.
Examples of the panoramic depth and reflectance images for each place category are shown in Fig.~\ref{fig:densempo_sample}.

\begin{table}[htb]
    \tbl{Distribution on Dense-MPO}
    {
        \centering
        \begin{tabular}{lllllllll}
            \toprule
            Category                                            & \multicolumn{7}{l}{Number of scans at each location} & Total                                                                                         \\ \cmidrule{2-8}
                        & {\tiny Set1} & {\tiny Set2} & {\tiny Set3} & {\tiny Set4} & {\tiny Set5} & {\tiny Set6} & {\tiny Set7} &     \\
            \midrule
            \scriptsize
            Coast       & 14           & 14           & 16           & 12           & 17           & 14           & 16           & 103 \\ \midrule
            Forest      & 16           & 16           & 17           & 18           & 16           & 16           & 17           & 116 \\ \midrule
            ParkingIn   & 16           & 16           & 13           & 15           & 17           & 13           & 13           & 105 \\ \midrule
            ParkingOut  & 15           & 17           & 16           & 15           & 15           & 14           & 16           & 108 \\ \midrule
            Residential & 14           & 16           & 14           & 15           & 16           & 15           & 16           & 106 \\ \midrule
            Urban       & 16           & 17           & 16           & 16           & 15           & 16           & 16           & 112 \\ \midrule
            \multicolumn{8}{r}{Total number of panoramic scans} & 650                                                                                                                                                  \\
            \bottomrule
        \end{tabular}
    }
    \label{table:densempo}.
\end{table}

\section{Convolutional Neural Networks for Panoramic Images}
\label{sec:models}
In this research, we treat the measured point clouds scattered on the 3D space and the corresponding reflectance values as panoramic images by projecting them onto the 2D plane.
This section describes CNN architectures and the training procedure to automatically learn feature representations from the panoramic images obtained from LiDARs.

\subsection{Data Preprocessing}
The panoramic depth and reflectance images given to our networks are converted from point clouds.
Each point on the scan is mapped into a 2D plane by cylindrical projection around the vertical axis.
On both depth and reflectance modalities, the obtained 2D maps are scaled by the max limit value of the LiDARs and fed into the networks.
To run our algorithms on the limited video memory of our GPU and to interpolate some defects on the images, we reduced the size of the panoramic images to $384\times32$ in the sparse set, by downsampling using bilinear interpolation.

\subsection{Models}

In this section, we first introduce uni-modal models for panoramic image recognition to verify the accuracy and the optimal architecture when either a depth map or a reflectance map is used. Then we introduce four types of multi-modal models that fuse both modalities.

\subsubsection{Uni-modal}
\label{sec:unimodal}
Our uni-modal CNN is based on the VGG-11 described in the paper by Simonyan \textit{et al.}~\cite{simonyan2014very}.
Here, we prepare a baseline model based on the VGG-11 to verify the effectiveness of our proposed architecture.
First, we shrank the last stack of fully-connected layers to fewer layers and units.
The shrinking reduced the large number of training parameters and resulted in fast convergence and higher performance.
Second, we applied batch normalization~\cite{ioffe2015batch} to the outputted feature maps of all convolution layers and all fully-connected layers except for the final layer.
The final architecture of the baseline is shown in Table~\ref{tab:arch}.
The performance was not improved even when using the deeper model in the paper~\cite{simonyan2014very} or skip-connections~\cite{he2016deep}.

To improve the performance of the baseline, we propose the following:

\begin{table}[t]
    \tbl{Baseline Architecture}
    {
        \centering
        \begin{tabular}{lllll}
            \toprule
            {Layer name}      & {Layer type}    & {Kernel}   & {Stride}   & {Parameters} \\
            \midrule
            \texttt{conv1}    & Convolution     & $3\times3$ & $1\times1$ & 64 filters   \\
            \texttt{pool1}    & Max Pooling     & $2\times2$ & $2\times2$ &              \\
            \midrule
            \texttt{conv2}    & Convolution     & $3\times3$ & $1\times1$ & 128 filters  \\
            \texttt{pool2}    & Max Pooling     & $2\times2$ & $2\times2$ &              \\
            \midrule
            \texttt{conv3\_1} & Convolution     & $3\times3$ & $1\times1$ & 256 filters  \\
            \texttt{conv3\_2} & Convolution     & $3\times3$ & $1\times1$ & 256 filters  \\
            \texttt{pool3}    & Max Pooling     & $2\times2$ & $2\times2$ &              \\
            \midrule
            \texttt{conv4\_1} & Convolution     & $3\times3$ & $1\times1$ & 512 filters  \\
            \texttt{conv4\_2} & Convolution     & $3\times3$ & $1\times1$ & 512 filters  \\
            \texttt{pool4}    & Max Pooling     & $2\times2$ & $2\times2$ &              \\
            \midrule
            \texttt{conv5\_1} & Convolution     & $3\times3$ & $1\times1$ & 512 filters  \\
            \texttt{conv5\_2} & Convolution     & $3\times3$ & $1\times1$ & 512 filters  \\
            \texttt{pool5}    & Max Pooling     & $2\times2$ & $2\times2$ &              \\
            \midrule
            \texttt{fc1}      & Fully-connected &            &            & 128 units    \\
            \texttt{drop}     & Dropout         &            &            &              \\
            \texttt{fc2}      & Fully-connected &            &            & 6 units      \\
            & Softmax         &                                        \\
            \bottomrule
        \end{tabular}
    }
    \label{tab:arch}
\end{table}

\subparagraph{Circular Convolution on the 2D Plane}
In a normal convolution layer, zero padding operation which fills the periphery of an incoming tensor with zero is commonly used to prevent reduction of an output size of feature maps.
In this paper, we propose a horizontal circular convolution (HCC) layer which extracts features while retaining the circular structure of the given panoramic images.
We replace all convolution layers of the baseline model with the HCC layer.
The HCC layer circulates the data flow in the horizontal direction, in forward padding processing and backward gradient calculation.

\subparagraph{Horizontally-Invariant Pooling}
The visual concepts on the panoramic images tend to move largely in the horizontal direction due to the yawing motion of the measurement vehicle and the installation angle of LiDAR.
We apply a row-wise max pooling (RWMP) layer  proposed by Shi \textit{et al.}~\cite{shi2015deeppano} before the first fully-connected layer (\texttt{fc1} in Table~\ref{tab:arch}).
The RWMP outputs the maximum value of each row of the input feature map and gives the invariance of horizontal translation to the CNNs processing a circular-structured image.

\subsubsection{Multi-modal}
\label{sec:multimodal}
\begin{figure}[t]
    \centering
    \includegraphics[width=\hsize]{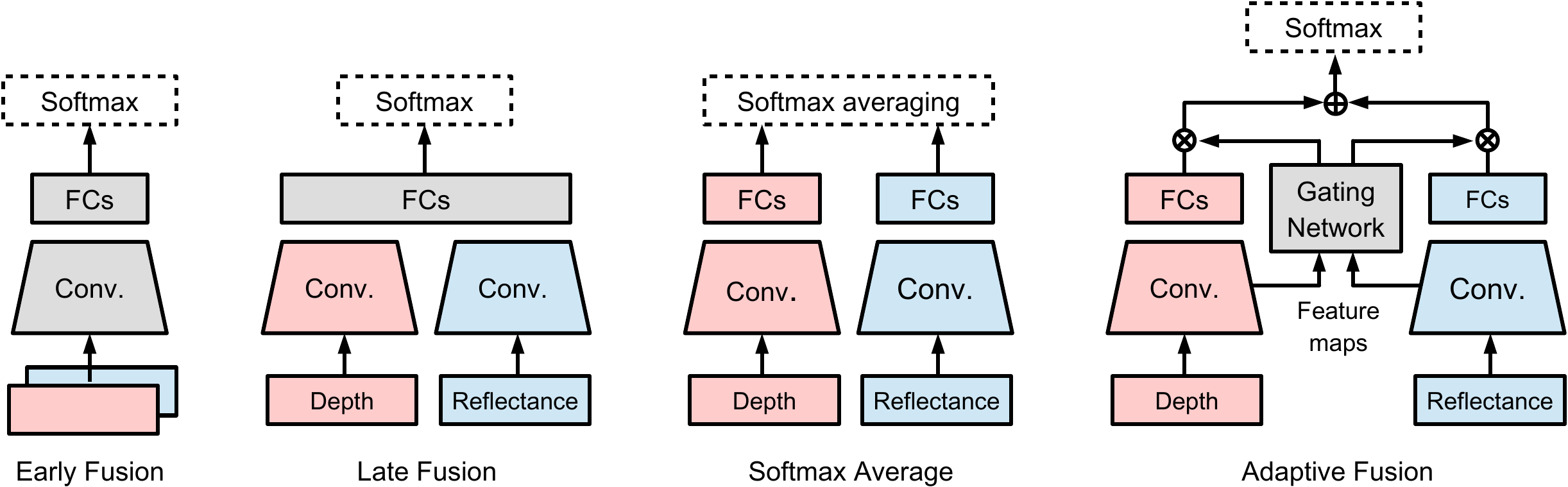}
    \caption{Architecture of multi-modal models}
    \label{fig:multi_arch}
\end{figure}

Multi-modal models receive both a depth map and corresponding reflectance map as input.
We herein introduce the four types of architectures for fusing the multiple modalities as depicted in Figure~\ref{fig:multi_arch}.
``FCs'' and ``Conv.'' denote fully-connected layers and convolution layers, respectively.
Let $\mathcal{D}=\{(X_n, y_n), n=1,2,\dots,N\}$ be the training samples, where $X_n$ is an input set of depth map $x_{nd}$ and reflectance map $x_{nr}$, and $y_n$ is a $C$-dimensional one-hot vector as ground-truth.

\subparagraph{Softmax Average}
The visual features of the depth map and the reflectance map are learned separately on the different models $f_d(\cdot)$ and $f_r(\cdot)$.
Each model is selected in terms of performance for the uni-modal case.
At testing, each of the trained models outputs probabilities of $C$ categories and a final classification result $c$ is decided from a category-wise averaged score $P(\bm{x}) \in \mathbb{R}^C$ as follows:
\begin{eqnarray}
    c = \argmax P(x)\\
    P(x) = \frac{f_d(x_d) + f_r(x_r)}{2}
\end{eqnarray}

\subparagraph{Adaptive Fusion}
The Adaptive Fusion model adaptively weights the estimated probabilities of the individually-trained models, by adopting an additional \textit{gating network} $g(\cdot)$ which estimates the certainty of each model from intermediate activations of the models.
Each model is selected in the same fashion as the Softmax Average.
This methodology was originally proposed by Mees \textit{et al.}~\cite{mees2016choosing} for a pedestrian detection task.
We chose the last convoluted feature \texttt{pool5} as the activations to input to the gating network.
At training, each uni-modal model is pre-trained separately and fixed, and then only the gating network is trained to maximize the weighted score of a correct label in $P(\bm{x})$.
\begin{eqnarray}
    P(\bm{x})=\underbrace{w_d f_d(x_d)}_{Depth} + \underbrace{w_r f_r(x_r)}_{Reflectance}\\
    (w_d, w_r)=g(r_d(x_d),r_r(x_r))
\end{eqnarray}
where $w_d$ and $w_r$ are the certainties estimated from the gating network $g(\cdot)$, and $r_d(\cdot)$ and $r_r(\cdot)$ are the intermediate activations of the depth and the reflectance models, respectively.

\subparagraph{Early Fusion}
The Early Fusion model takes two-channel input that is a simple stack of the modalities.
Local co-occurrence features can be learned on the pixel level since the information of both modalities is merged at the first convolution layer of the unified network.
Except for the number of input channels, the layer structure is the same as the uni-modal baseline.
All parameters of the input through the output are trained end-to-end.

\subparagraph{Late Fusion}
The Late Fusion model has two convolution streams to extract the modal-wise features separately and share fully-connected layers to output a final categorical distribution.
The modal-wise features are abstracted spatial information and finally are combined in the first fully-connected layer to compute more semantic co-occurrence features.
Up to the last convolution layers, the architectures are same as in the uni-modal cases.
The fully-connected layers are also the same but shared between the modalities.

\subsection{Data Augmentation}
We augment the training data by applying two types of random transformations to the original image set. First, the input image is horizontally flipped. Second, we perform a random circular shift on the image in the horizontal direction; this is equivalent to rotating panoramic images in the yawing direction. The number of shifted pixels is randomly selected from zero to the image width.

\subsection{Training}
The weights of the network are trained by a backpropagation algorithm using stochastic gradient descent (SGD) with a fixed learning rate of $10^{-4}$ and a momentum of 0.9.
In each propagation, a mini-batch of images is fed into the network to estimate a categorical distribution of each image.
We empirically set the training mini-batch size to 64.
Furthermore, we perform $L_2$-regularization on the network by weight decay~\cite{moody1995simple} to mitigate the over-fitting risk.
The coefficient of the regularization term is fixed at $5\times10^{-4}$.
Dropout regularization of 50\% is added to the output of the first fully-connected layer; this regularization allows us to explore generalized weights while avoiding over-fitting.
\section{Experiments}
\label{sec:experiment}
The experiments are composed of two parts, \textit{uni-modal} and \textit{multi-modal}.
In the uni-modal experiment, either a depth map or a reflectance map of Sparse MPO is given to the proposed models described in Section~\ref{sec:unimodal}.
Subsequently, we conduct the multi-modal experiment with the models described in Section~\ref{sec:multimodal} by using the uni-modal results.

Our algorithms are implemented with a deep learning framework PyTorch~\cite{paszke2017automatic}, and the training and evaluation are performed on NVIDIA Geforce GTX Titan X.

\subsection{Settings and Evaluation Metric}
Our validation uses a $k$-fold cross-validation approach in which we use one set in each category for testing and the rest for training. We repeat this process $k$ times with different disjoint training and tests sets.
In a training phase, we extract a validation set from the training sets to monitor overfitting.
When the validation loss does not improve over 10 epochs, i.e. the validation overfits or plateaus on the training sets, the training is stopped early and then we use the parameters for testing.
Finally, we report a mean accuracy over the $k$-time testing.
The numbers of folds $k$ are 10 for the sparse set, according to the scanning groups of the datasets.

\subsection{Uni-modal Performance on Sparse MPO}
In this section, we report the performance of the following four types of CNNs on two types of modalities, depth and reflectance: baseline VGG, VGG in which either HCC layers or RWMP layer is applied (VGG+HCC, VGG+RWMP), and VGG in which both HCC layers and RWMP layer are applied (VGG+RWMP+HCC).
Furthermore, we compare these results to those by traditional \textit{hand-engineered} approaches and another CNN variant, ResNet~\cite{he2016deep}.
We chose the 20-layer ResNet proposed in the CIFAR-10 experiment~\cite{he2016deep} because the image size is similar to ours.
The performance of all the approaches is shown in Table~\ref{table:unimodal}.

\subsubsection{Comparison of CNN models}
\label{sec:unimodal_comp_vgg}
\begin{figure}
    \centering
    \subfigure[Depth: VGG]{
        \includegraphics[width=0.45\hsize]{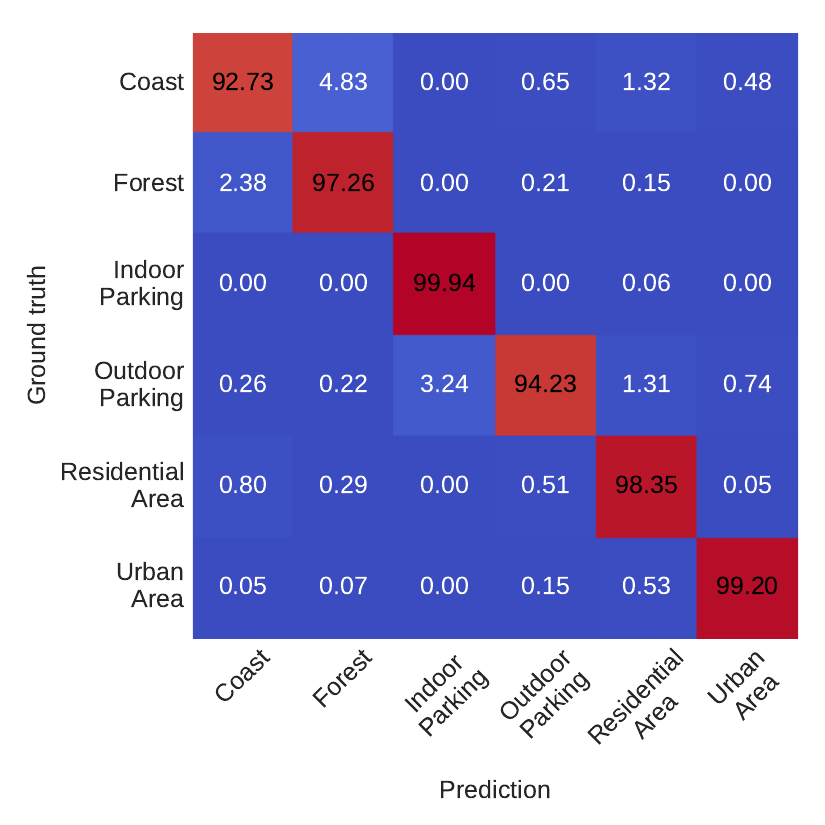}
    }
    \hfill
    \subfigure[Reflectance: VGG+RWMP+HCC]{
        \includegraphics[width=0.45\hsize]{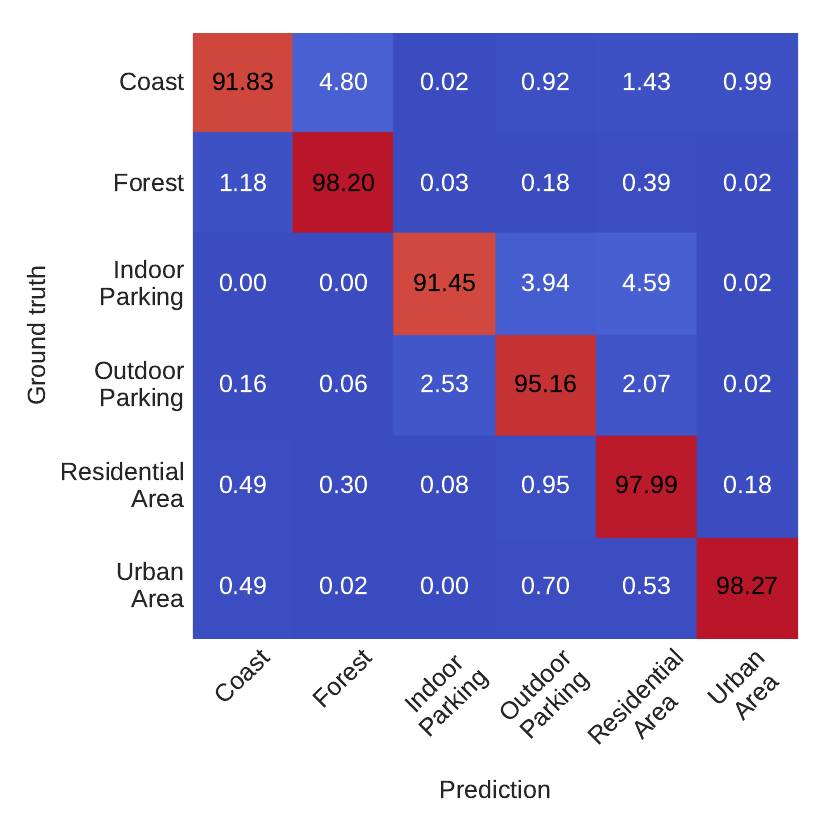}
    }
    \caption{Confusion matrix of best models on a depth and reflectance inputs.}
    \label{fig:confusion_matrix}
\end{figure}

As shown in Table~\ref{table:unimodal}, for reflectance input, the VGG+RWMP+HCC achieved the best accuracy of 95.92\% in total.
On the other hand, for depth input, the baseline VGG without any modification achieved the best accuracy of 97.18\% in total.
Focusing on the results for each category, we can see that the best reflectance model shows higher accuracy in the \textit{Forest} and \textit{ParkingOut} categories than the best depth model.
It can be seen that the accuracy can be improved by combining both modalities.
Confusion matrices of the best models with depth and reflectance are shown in Fig.~\ref{fig:confusion_matrix}.
The models tend to have more errors between \textit{Coast} and \textit{Forest} categories.
One of the reasons is that the images in \textit{Coast} category are unique in that the data is lacked in the sky and sea areas, and are similar to \textit{Forest} category in the opposite wooded areas. It can be considered that the model would misclassify the \textit{Coast} images into \textit{Forest} if its unique areas are widely covered by crossing cars or trees.
In Section~\ref{sec:uni_qa}, we discuss further details about the effects of the proposed layers.

\subsubsection{Comparison to traditional approaches}
We compare the categorization results with the three traditional approaches, Spin-Image~\cite{John1997}, GIST~\cite{oliva2001modeling}, and Local Binary Patterns (LBP)~\cite{Ojala2002}, to extract hand-engineered features from a given image.
Spin-Image is one of the popular technique for surface matching and 3D object recognition tasks.
In this paper we obtains a single Spin-Image representation from an original point cloud in a scanner-oriented viewpoint in the cylindrical coordinate system.
GIST~\cite{oliva2001modeling} was proposed specifically for scene categorization task.
Some recent studies~\cite{Song:2015js, xiao2010sun} on the scene categorization used the GIST features for color and depth images.
Although LBP was proposed originally for texture classification by Ojala \textit{et al.}~\cite{Ojala2002}, it is reported that LBP is also effective for indoor place categorization~\cite{Mozos2013,Mozos2012}.
These features are given to a SVM with a radial basis function (RBF) kernel to categorize places.
The parameters of the kernel are selected by a grid search.
In all categories, our data-driven approaches using CNNs outperforms the traditional methods above.

\begin{table*}[t]
    \tbl{Classification accuracy [\%] on uni-modal input.}
    {
        \centering
        \begin{tabular}{lllcccccccc}
            \toprule
            Input       & Approach                                &   & Coast          & Forest         & In. P.         & Out. P.        & Res.           & Urban          &   & Total          \\
            \midrule
            Depth       & Spin-Image~\cite{John1997} + RBF-SVM    &   & 65.60          & 86.30          & 81.84          & 86.26          & 82.95          & 64.31          &   & 79.23          \\
                        & GIST~\cite{oliva2001modeling} + RBF-SVM &   & 75.42          & 91.52          & 82.72          & 86.72          & 86.54          & 81.16          &   & 84.53          \\
                        & LBP~\cite{Ojala2002} + RBF-SVM          &   & 84.25          & 94.93          & 96.41          & 86.86          & 94.58          & 92.71          &   & 92.00          \\
            \cmidrule{2-2}
                        & ResNet~\cite{he2016deep}                &   & 90.65          & 94.25          & 97.67          & \textbf{95.00} & 97.68          & 97.79          &   & 95.66          \\
                        & \textbf{VGG}                            &   & 92.73          & 97.26          & \textbf{99.94} & 94.23          & \textbf{98.35} & \textbf{99.20} &   & \textbf{97.18} \\
                        & VGG + RWMP                              &   & \textbf{93.74} & \textbf{97.53} & 98.98          & 94.45          & \textbf{98.35} & 98.35          &   & 97.11          \\
                        & VGG + HCC                               &   & 92.98          & 96.54          & 99.54          & 93.99          & 98.23          & 98.81          &   & 96.89          \\
                        & VGG + RWMP + HCC                        &   & 93.03          & 96.80          & 98.85          & 94.29          & 98.32          & 98.86          &   & 96.92          \\
            \midrule
            Reflectance & GIST~\cite{oliva2001modeling} + RBF-SVM &   & 68.85          & 91.65          & 74.07          & 81.73          & 79.88          & 75.03          &   & 79.15          \\
                        & LBP~\cite{Ojala2002} + RBF-SVM          &   & 76.66          & 95.67          & 80.18          & 92.74          & 88.44          & 79.53          &   & 86.19          \\
            \cmidrule{2-2}
                        & ResNet~\cite{he2016deep}                &   & 90.30          & 96.62          & \textbf{94.60} & 93.63          & 96.13          & 96.09          &   & 94.83          \\
                        & VGG                                     &   & 90.58          & 97.99          & 92.37          & 93.89          & 96.40          & 95.06          &   & 94.75          \\
                        & VGG + RWMP                              &   & 91.13          & 97.63          & 93.29          & 94.67          & 97.93          & 97.37          &   & 95.74          \\
                        & VGG + HCC                               &   & 91.00          & \textbf{98.38} & 94.39          & 94.13          & 97.58          & 95.01          &   & 95.45          \\
                        & \textbf{VGG + RWMP + HCC}               &   & \textbf{91.83} & 98.20          & 91.45          & \textbf{95.16} & \textbf{97.99} & \textbf{98.27} &   & \textbf{95.92} \\
            \bottomrule
        \end{tabular}
    }
    \label{table:unimodal}
\end{table*}

\subsubsection{Robustness to a rotation on the images}
We rotated the input image horizontally and examined the effects on the classification accuracy. The results are shown in Fig.~\ref{fig:rotating}. For both depth and reflectance modalities, it can be seen that the accuracy of the baseline VGG and VGG+RWMP models is slightly dropped when the angle of rotation is around \ang{90} or \ang{270}.
A further analysis on this point is discussed in Section~\ref{sec:uni_qa}.
On the other hand, the RWMP and HCC modification contributes to rotational invariance for both input modalities.
The cyclic ripples seen in the accuracy depend on a size of a receptive field modified by RWMP and HCC.
Feature maps given to RWMP have a horizontal resolution of 12 units at each row, thus the receptive field on the input image has a \ang{30} ($\ang{360}/12$) horizontal periodicity.

\subsubsection{Qualitative Analysis}
\label{sec:uni_qa}
\begin{figure*}
    \centering
    \subfigure[Depth input]
    {
        \includegraphics[width=0.48\hsize]{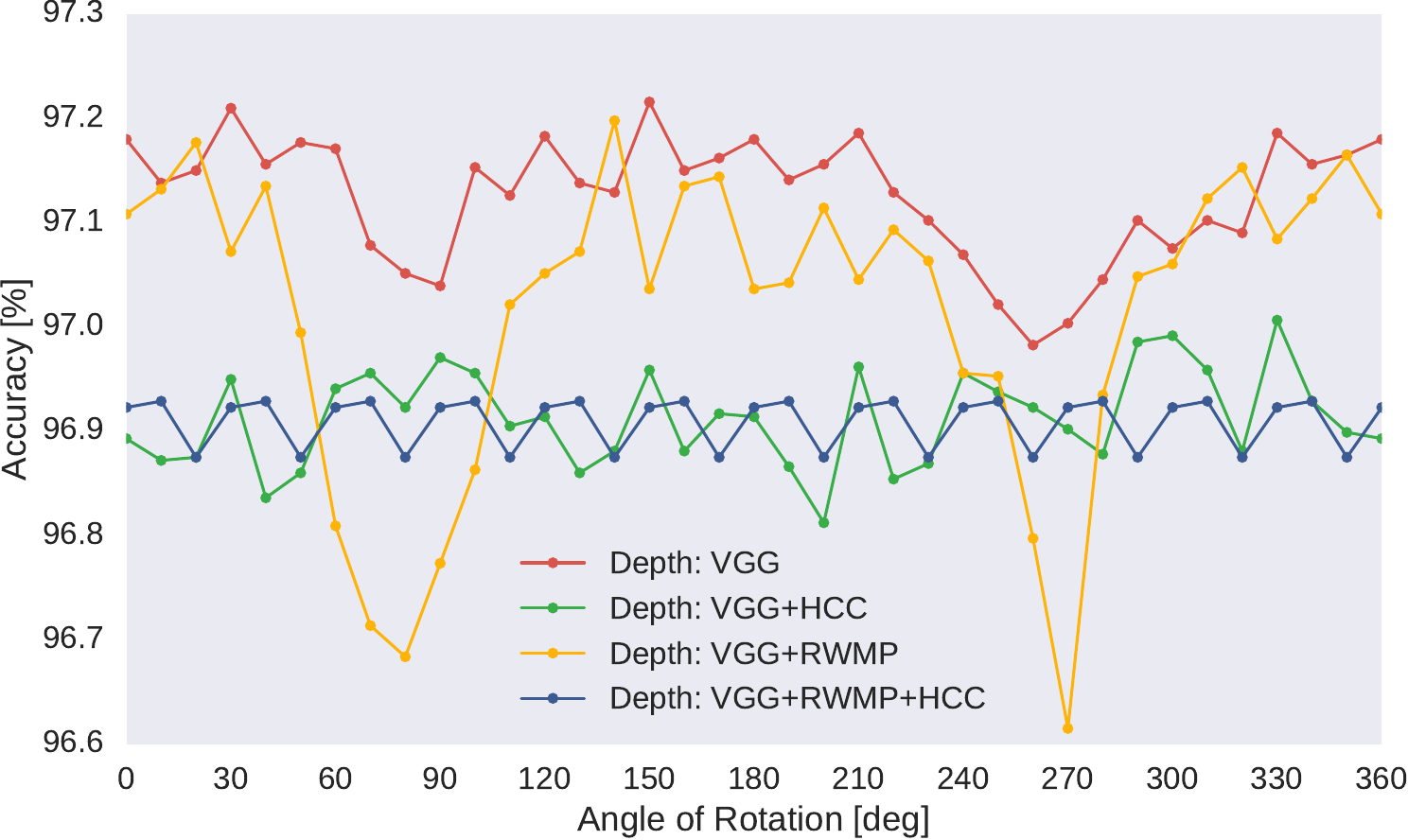}
    }
    \hfill
    \subfigure[Reflectance input]
    {
        \includegraphics[width=0.48\hsize]{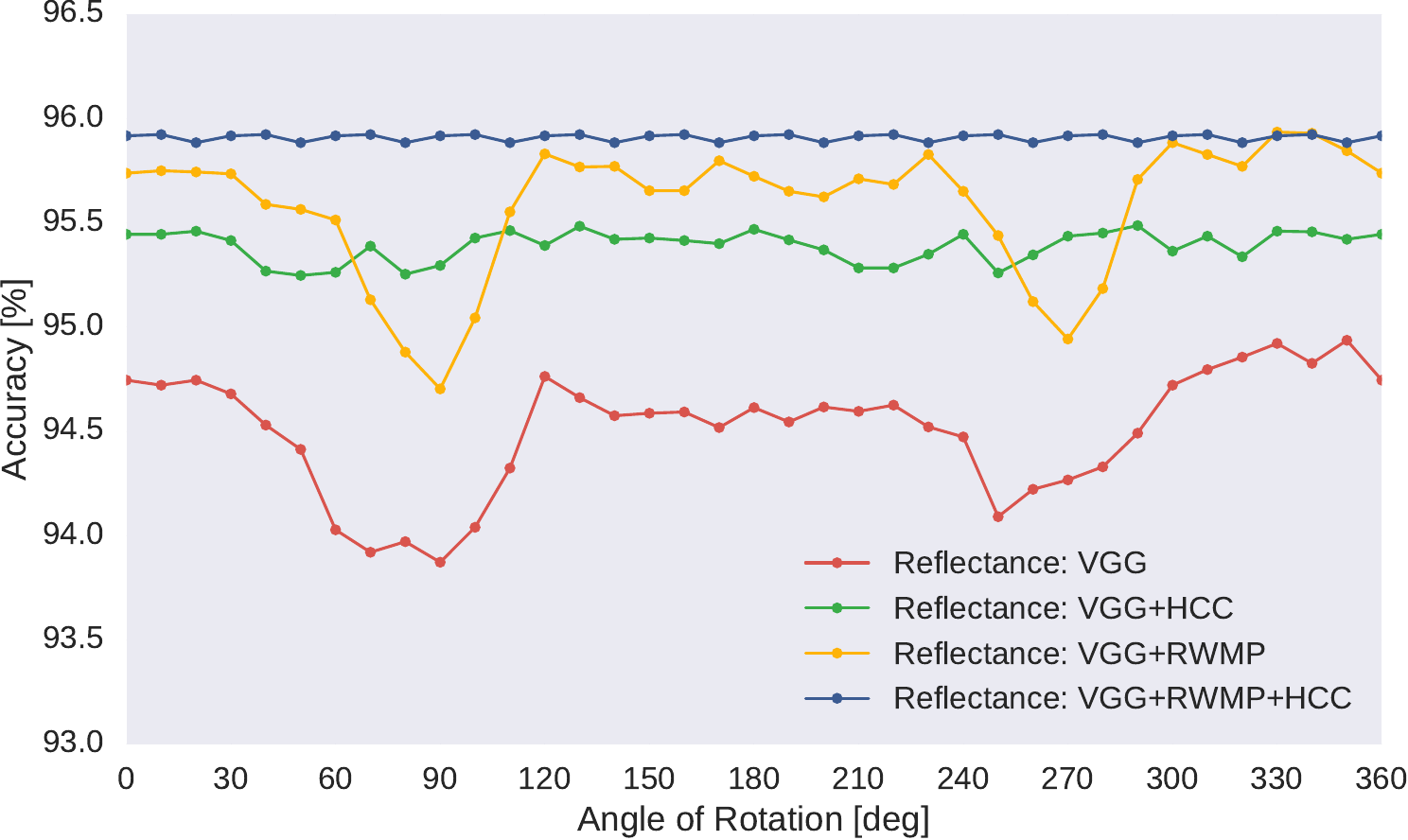}
    }
    \caption{Horizontal rotation of an input image vs. accuracy.}
    \label{fig:rotating}
\end{figure*}

\begin{figure}
    \centering
    \begin{minipage}[c]{0.05\hsize}\centering{\footnotesize Angle}\end{minipage}
    \begin{minipage}[c]{0.3\hsize}\centering{\footnotesize \textit{Coast} example}\end{minipage} \hfill
    \begin{minipage}[c]{0.3\hsize}\end{minipage} \hfill
    \begin{minipage}[c]{0.3\hsize}\end{minipage}
    \\
    \vspace{1mm}
    \begin{minipage}[c]{0.05\hsize}\centering {\footnotesize \ang{0}}\end{minipage}
    \begin{minipage}[c]{0.3\hsize}\includegraphics[width=\hsize]{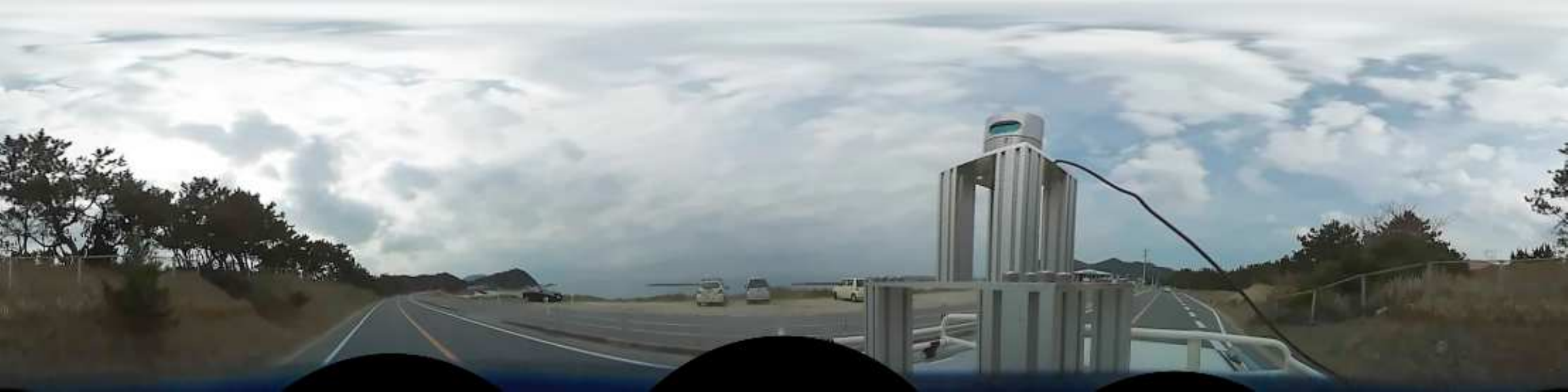}\end{minipage} \hfill
    \begin{minipage}[c]{0.3\hsize}\end{minipage} \hfill
    \begin{minipage}[c]{0.3\hsize}\end{minipage}
    \vspace{1mm}
    \\
    \begin{minipage}[c]{0.05\hsize}\centering{\footnotesize Angle}\end{minipage}
    \begin{minipage}[c]{0.3\hsize}\centering{\footnotesize Depth input}\end{minipage} \hfill
    \begin{minipage}[c]{0.3\hsize}\centering{\footnotesize Grad-CAM: VGG}\end{minipage} \hfill
    \begin{minipage}[c]{0.3\hsize}\centering{\footnotesize Grad-CAM: VGG+RWMP+HCC}\end{minipage}
    \\
    \begin{minipage}[c]{0.05\hsize}\centering {\footnotesize \ang{0}}\end{minipage}
    \begin{minipage}[c]{0.3\hsize}\includegraphics[width=\hsize]{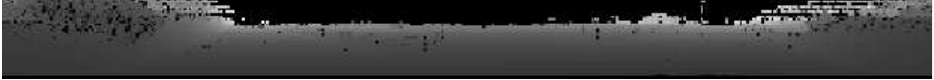}\end{minipage} \hfill
    \begin{minipage}[c]{0.3\hsize}\includegraphics[width=\hsize]{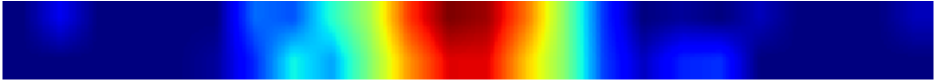}\end{minipage} \hfill
    \begin{minipage}[c]{0.3\hsize}\includegraphics[width=\hsize]{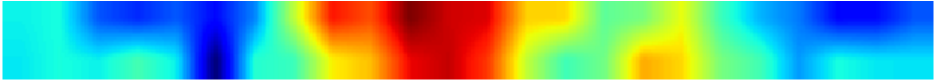}\end{minipage}
    \\
    \begin{minipage}[c]{0.05\hsize}\centering {\footnotesize \ang{90}}\end{minipage}
    \begin{minipage}[c]{0.3\hsize}\includegraphics[width=\hsize]{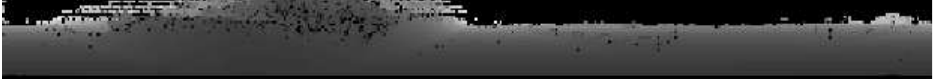}\end{minipage} \hfill
    \begin{minipage}[c]{0.3\hsize}\includegraphics[width=\hsize]{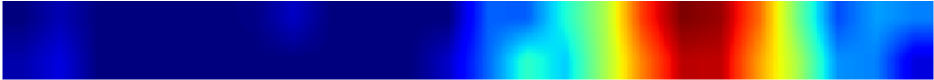}\end{minipage} \hfill
    \begin{minipage}[c]{0.3\hsize}\includegraphics[width=\hsize]{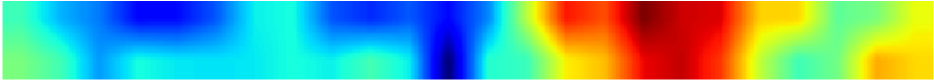}\end{minipage}
    \\
    \begin{minipage}[c]{0.05\hsize}\centering {\footnotesize \ang{180}}\end{minipage}
    \begin{minipage}[c]{0.3\hsize}\includegraphics[width=\hsize]{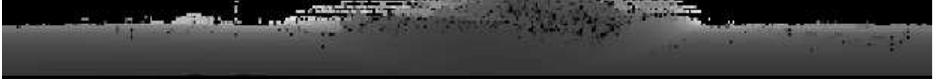}\end{minipage} \hfill
    \begin{minipage}[c]{0.3\hsize}\includegraphics[width=\hsize]{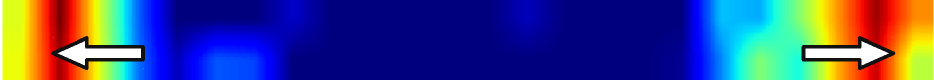}\end{minipage} \hfill
    \begin{minipage}[c]{0.3\hsize}\includegraphics[width=\hsize]{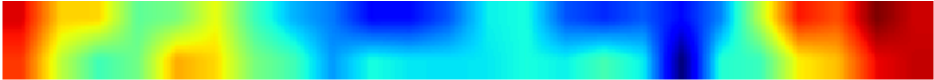}\end{minipage}
    \\
    \begin{minipage}[c]{0.05\hsize}\centering {\footnotesize \ang{270}}\end{minipage}
    \begin{minipage}[c]{0.3\hsize}\includegraphics[width=\hsize]{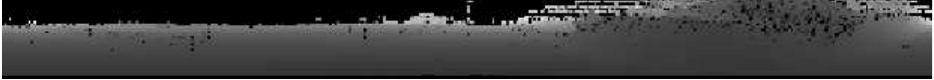}\end{minipage} \hfill
    \begin{minipage}[c]{0.3\hsize}\includegraphics[width=\hsize]{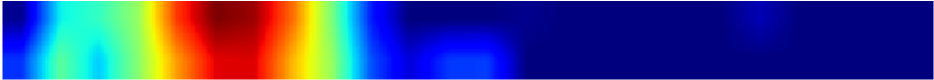}\end{minipage} \hfill
    \begin{minipage}[c]{0.3\hsize}\includegraphics[width=\hsize]{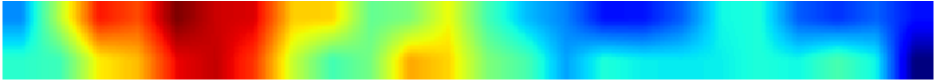}\end{minipage}
    \\
    \begin{minipage}[c]{0.05\hsize}\centering{\footnotesize Angle}\end{minipage}
    \begin{minipage}[c]{0.3\hsize}\centering{\footnotesize Reflectance input}\end{minipage} \hfill
    \begin{minipage}[c]{0.3\hsize}\centering{\footnotesize Grad-CAM: VGG}\end{minipage} \hfill
    \begin{minipage}[c]{0.3\hsize}\centering{\footnotesize Grad-CAM: VGG+RWMP+HCC}\end{minipage}
    \\
    \begin{minipage}[c]{0.05\hsize}\centering {\footnotesize \ang{0}}\end{minipage}
    \begin{minipage}[c]{0.3\hsize}\includegraphics[width=\hsize]{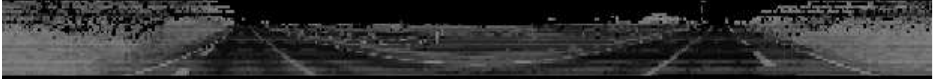}\end{minipage} \hfill
    \begin{minipage}[c]{0.3\hsize}\includegraphics[width=\hsize]{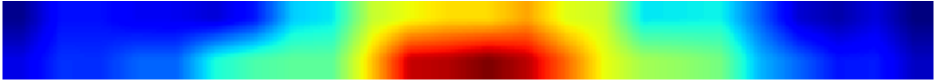}\end{minipage} \hfill
    \begin{minipage}[c]{0.3\hsize}\includegraphics[width=\hsize]{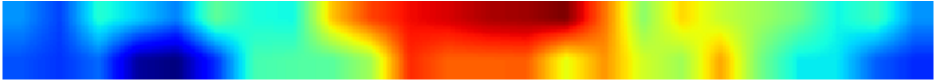}\end{minipage}
    \\
    \begin{minipage}[c]{0.05\hsize}\centering {\footnotesize \ang{90}}\end{minipage}
    \begin{minipage}[c]{0.3\hsize}\includegraphics[width=\hsize]{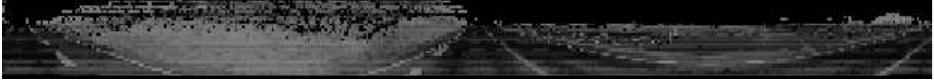}\end{minipage} \hfill
    \begin{minipage}[c]{0.3\hsize}\includegraphics[width=\hsize]{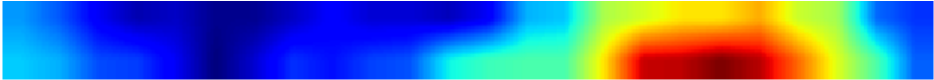}\end{minipage} \hfill
    \begin{minipage}[c]{0.3\hsize}\includegraphics[width=\hsize]{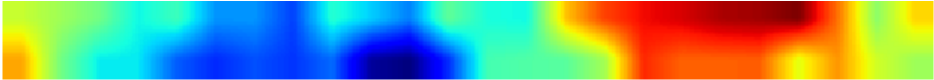}\end{minipage}
    \\
    \begin{minipage}[c]{0.05\hsize}\centering {\footnotesize \ang{180}}\end{minipage}
    \begin{minipage}[c]{0.3\hsize}\includegraphics[width=\hsize]{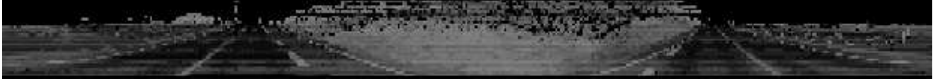}\end{minipage} \hfill
    \begin{minipage}[c]{0.3\hsize}\includegraphics[width=\hsize]{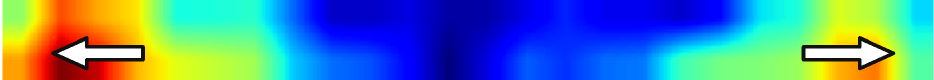}\end{minipage} \hfill
    \begin{minipage}[c]{0.3\hsize}\includegraphics[width=\hsize]{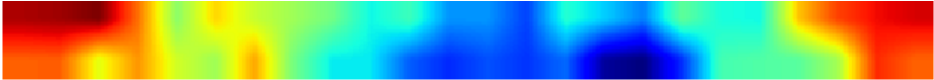}\end{minipage}
    \\
    \begin{minipage}[c]{0.05\hsize}\centering {\footnotesize \ang{270}}\end{minipage}
    \begin{minipage}[c]{0.3\hsize}\includegraphics[width=\hsize]{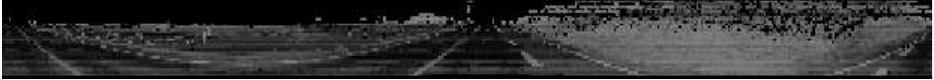}\end{minipage} \hfill
    \begin{minipage}[c]{0.3\hsize}\includegraphics[width=\hsize]{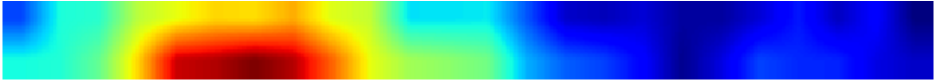}\end{minipage} \hfill
    \begin{minipage}[c]{0.3\hsize}\includegraphics[width=\hsize]{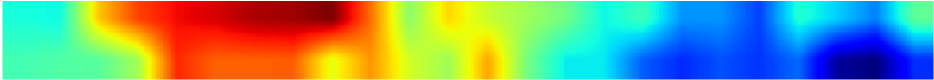}\end{minipage}
    %
    %
    %
    \\
    \begin{minipage}[c]{0.05\hsize}\centering{\footnotesize Angle}\end{minipage}
    \begin{minipage}[c]{0.3\hsize}\centering{\footnotesize \textit{Residential} example}\end{minipage} \hfill
    \begin{minipage}[c]{0.3\hsize}\end{minipage} \hfill
    \begin{minipage}[c]{0.3\hsize}\end{minipage}
    \\
    \vspace{1mm}
    \begin{minipage}[c]{0.05\hsize}\centering {\footnotesize \ang{0}}\end{minipage}
    \begin{minipage}[c]{0.3\hsize}\includegraphics[width=\hsize]{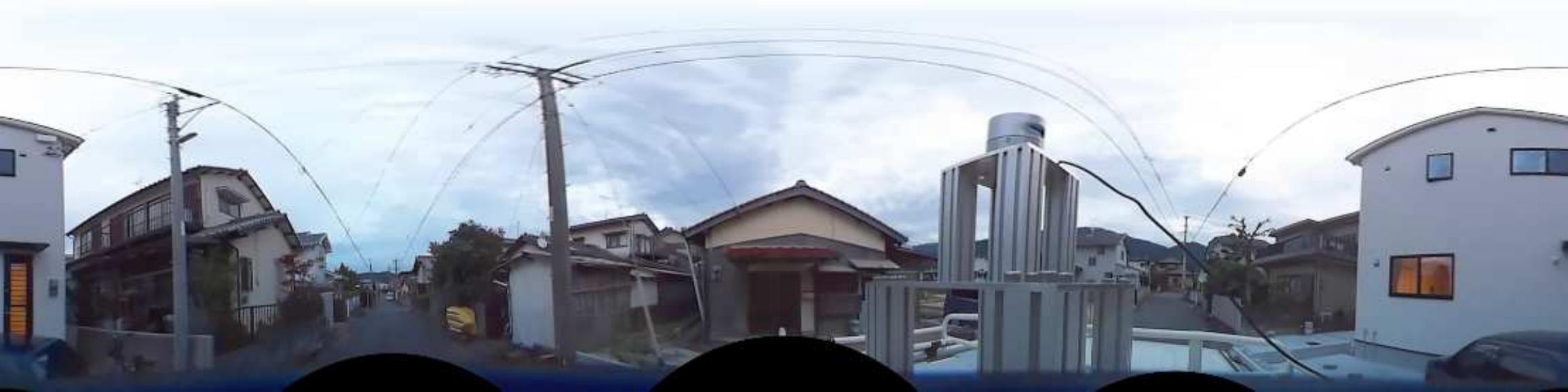}\end{minipage} \hfill
    \begin{minipage}[c]{0.3\hsize}\end{minipage} \hfill
    \begin{minipage}[c]{0.3\hsize}\end{minipage}
    \vspace{1mm}
    \\
    \begin{minipage}[c]{0.05\hsize}\centering{\footnotesize Angle}\end{minipage}
    \begin{minipage}[c]{0.3\hsize}\centering{\footnotesize Depth input}\end{minipage} \hfill
    \begin{minipage}[c]{0.3\hsize}\centering{\footnotesize Grad-CAM: VGG}\end{minipage} \hfill
    \begin{minipage}[c]{0.3\hsize}\centering{\footnotesize Grad-CAM: VGG+RWMP+HCC}\end{minipage}
    \\
    \begin{minipage}[c]{0.05\hsize}\centering {\footnotesize \ang{0}}\end{minipage}
    \begin{minipage}[c]{0.3\hsize}\includegraphics[width=\hsize]{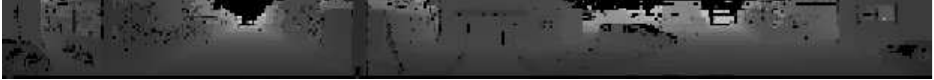}\end{minipage} \hfill
    \begin{minipage}[c]{0.3\hsize}\includegraphics[width=\hsize]{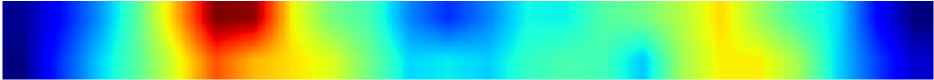}\end{minipage} \hfill
    \begin{minipage}[c]{0.3\hsize}\includegraphics[width=\hsize]{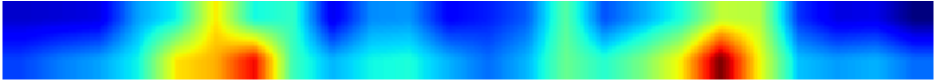}\end{minipage}
    \\
    \begin{minipage}[c]{0.05\hsize}\centering {\footnotesize \ang{90}}\end{minipage}
    \begin{minipage}[c]{0.3\hsize}\includegraphics[width=\hsize]{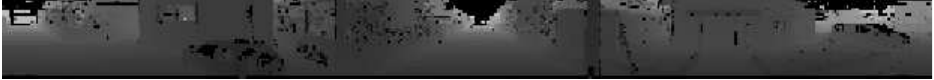}\end{minipage} \hfill
    \begin{minipage}[c]{0.3\hsize}\includegraphics[width=\hsize]{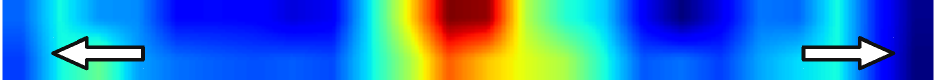}\end{minipage} \hfill
    \begin{minipage}[c]{0.3\hsize}\includegraphics[width=\hsize]{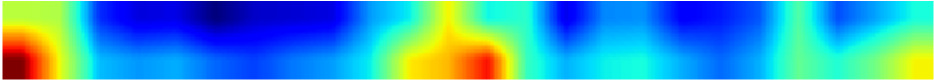}\end{minipage}
    \\
    \begin{minipage}[c]{0.05\hsize}\centering {\footnotesize \ang{180}}\end{minipage}
    \begin{minipage}[c]{0.3\hsize}\includegraphics[width=\hsize]{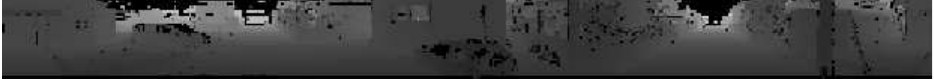}\end{minipage} \hfill
    \begin{minipage}[c]{0.3\hsize}\includegraphics[width=\hsize]{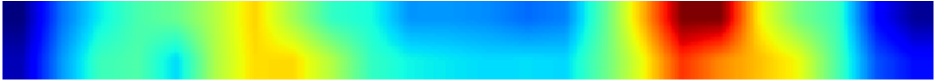}\end{minipage} \hfill
    \begin{minipage}[c]{0.3\hsize}\includegraphics[width=\hsize]{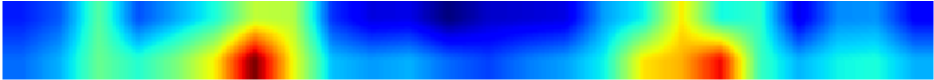}\end{minipage}
    \\
    \begin{minipage}[c]{0.05\hsize}\centering {\footnotesize \ang{270}}\end{minipage}
    \begin{minipage}[c]{0.3\hsize}\includegraphics[width=\hsize]{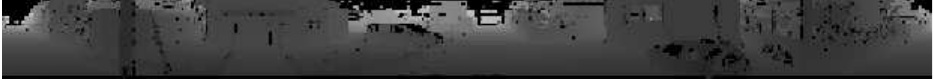}\end{minipage} \hfill
    \begin{minipage}[c]{0.3\hsize}\includegraphics[width=\hsize]{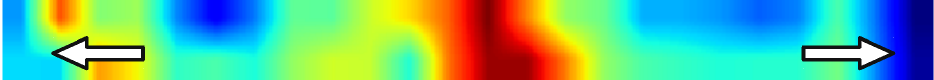}\end{minipage} \hfill
    \begin{minipage}[c]{0.3\hsize}\includegraphics[width=\hsize]{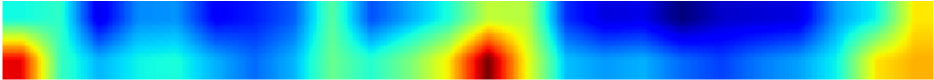}\end{minipage}
    \\
    \begin{minipage}[c]{0.05\hsize}\centering{\footnotesize Angle}\end{minipage}
    \begin{minipage}[c]{0.3\hsize}\centering{\footnotesize Reflectance input}\end{minipage} \hfill
    \begin{minipage}[c]{0.3\hsize}\centering{\footnotesize Grad-CAM: VGG}\end{minipage} \hfill
    \begin{minipage}[c]{0.3\hsize}\centering{\footnotesize Grad-CAM: VGG+RWMP+HCC}\end{minipage}
    \\
    \begin{minipage}[c]{0.05\hsize}\centering {\footnotesize \ang{0}}\end{minipage}
    \begin{minipage}[c]{0.3\hsize}\includegraphics[width=\hsize]{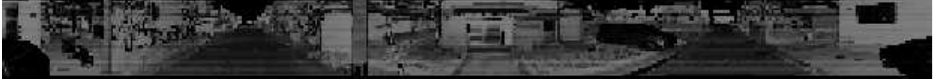}\end{minipage} \hfill
    \begin{minipage}[c]{0.3\hsize}\includegraphics[width=\hsize]{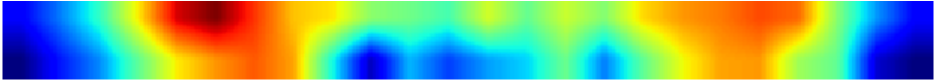}\end{minipage} \hfill
    \begin{minipage}[c]{0.3\hsize}\includegraphics[width=\hsize]{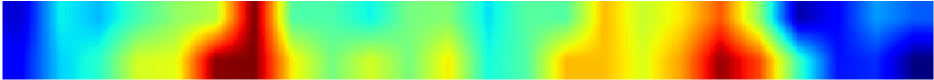}\end{minipage}
    \\
    \begin{minipage}[c]{0.05\hsize}\centering {\footnotesize \ang{90}}\end{minipage}
    \begin{minipage}[c]{0.3\hsize}\includegraphics[width=\hsize]{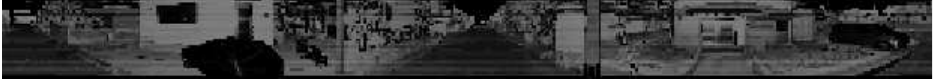}\end{minipage} \hfill
    \begin{minipage}[c]{0.3\hsize}\includegraphics[width=\hsize]{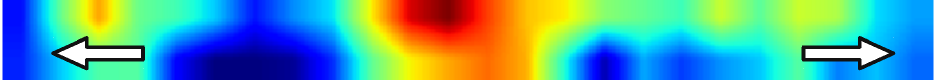}\end{minipage} \hfill
    \begin{minipage}[c]{0.3\hsize}\includegraphics[width=\hsize]{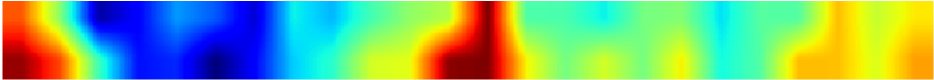}\end{minipage}
    \\
    \begin{minipage}[c]{0.05\hsize}\centering {\footnotesize \ang{180}}\end{minipage}
    \begin{minipage}[c]{0.3\hsize}\includegraphics[width=\hsize]{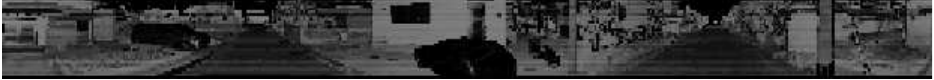}\end{minipage} \hfill
    \begin{minipage}[c]{0.3\hsize}\includegraphics[width=\hsize]{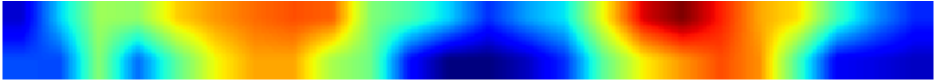}\end{minipage} \hfill
    \begin{minipage}[c]{0.3\hsize}\includegraphics[width=\hsize]{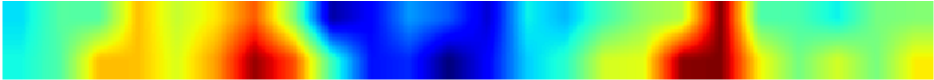}\end{minipage}
    \\
    \begin{minipage}[c]{0.05\hsize}\centering {\footnotesize \ang{270}}\end{minipage}
    \begin{minipage}[c]{0.3\hsize}\includegraphics[width=\hsize]{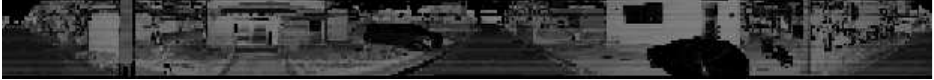}\end{minipage} \hfill
    \begin{minipage}[c]{0.3\hsize}\includegraphics[width=\hsize]{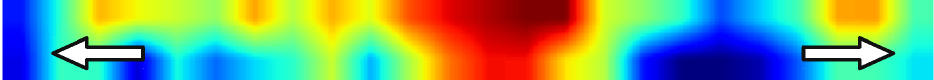}\end{minipage} \hfill
    \begin{minipage}[c]{0.3\hsize}\includegraphics[width=\hsize]{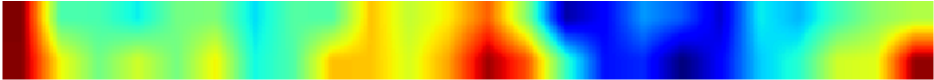}\end{minipage}
    \caption{Visualization of contributing features given rotated images, by Grad-CAM~\cite{Selvaraju_2017_ICCV}.}
    \label{fig:gradcam_rotating}
\end{figure}

To verify the way that the models solved the problem, we visualized feature spaces.
We visualize how the models process the images and predict the category by using Grad-CAM~\cite{Selvaraju_2017_ICCV}.
Grad-CAM enables us to see discriminative regions that strongly influence a specific category, by weighting feature maps of a certain layer with gradients derived from a score of the category.
We chose the \texttt{pool5} layer for extracting the feature maps.
Like in the previous section, we rotated an input image horizontally, and visualize regions contributing to the correct classification by Grad-CAM.
In Fig.~\ref{fig:gradcam_rotating}, we show results of two types of images.
One of them is a set of a depth and reflectance image from the \textit{Coast} category, which is strongly activated at the center region (lateral direction of the measurement vehicle).
Another one is a set from the \textit{Residential} category, which is activated at the two regions which are front and back areas of the vehicle.
It can be seen that all the models adaptively attend the discriminative regions according to the rotation.
However, in the baseline VGG model, the moving activated regions are attenuated or split at both ends (white arrows in Fig.~\ref{fig:gradcam_rotating}).
We consider that the accuracy drops in Fig.~\ref{fig:rotating} are due to the images like the \textit{Residential} example.

\begin{figure}
    \subfigure[\textit{Coast} on depth ]
    {
        \begin{tabular}{cc}
            \includegraphics[width=0.45\hsize]{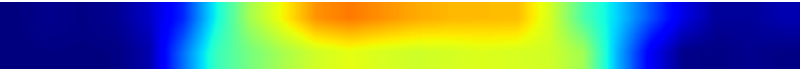}      \\
            \includegraphics[width=0.45\hsize]{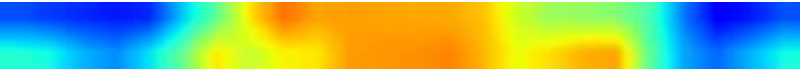}
        \end{tabular}
    }
    \hspace{5pt}
    \subfigure[\textit{Coast} on reflectance]
    {
        \begin{tabular}{cc}
            \includegraphics[width=0.45\hsize]{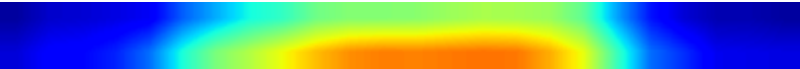}      \\
            \includegraphics[width=0.45\hsize]{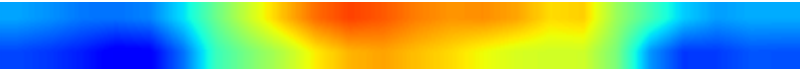}
        \end{tabular}
    }
    \\
    \subfigure[\textit{Forest} on depth ]
    {
        \begin{tabular}{cc}
            \includegraphics[width=0.45\hsize]{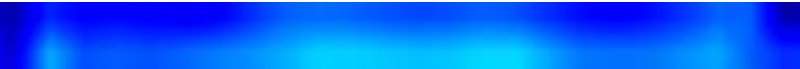}      \\
            \includegraphics[width=0.45\hsize]{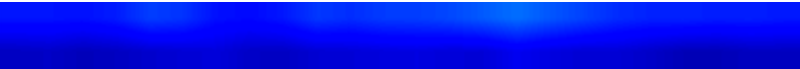}
        \end{tabular}
    }
    \hspace{5pt}
    \subfigure[\textit{Forest} on reflectance]
    {
        \begin{tabular}{cc}
            \includegraphics[width=0.45\hsize]{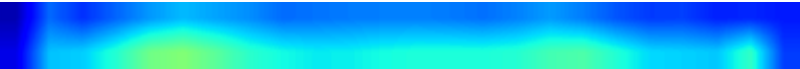}      \\
            \includegraphics[width=0.45\hsize]{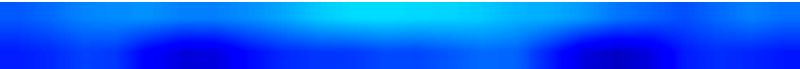}
        \end{tabular}
    }
    \\
    \subfigure[\textit{ParkingIn} on depth ]
    {
        \begin{tabular}{cc}
            \includegraphics[width=0.45\hsize]{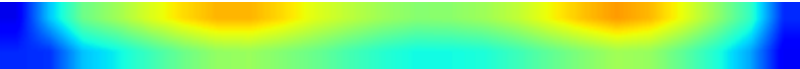}      \\
            \includegraphics[width=0.45\hsize]{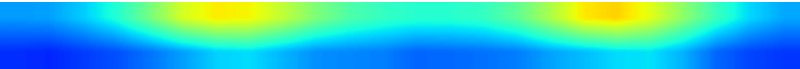}
        \end{tabular}
    }
    \hspace{5pt}
    \subfigure[\textit{ParkingIn} on reflectance]
    {
        \begin{tabular}{cc}
            \includegraphics[width=0.45\hsize]{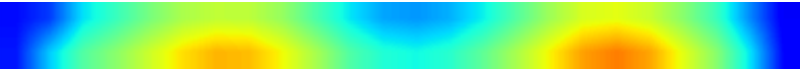}      \\
            \includegraphics[width=0.45\hsize]{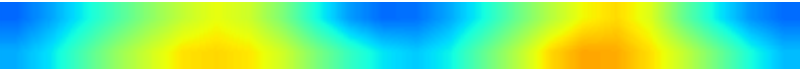}
        \end{tabular}
    }
    \\
    \subfigure[\textit{ParkingOut} on depth ]
    {
        \begin{tabular}{cc}
            \includegraphics[width=0.45\hsize]{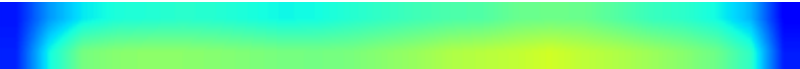}      \\
            \includegraphics[width=0.45\hsize]{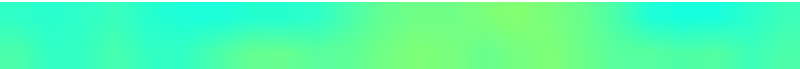}
        \end{tabular}
    }
    \hspace{5pt}
    \subfigure[\textit{ParkingOut} on reflectance]
    {
        \begin{tabular}{cc}
            \includegraphics[width=0.45\hsize]{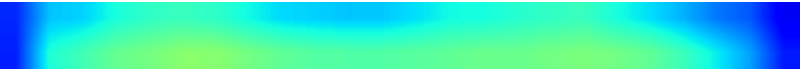}      \\
            \includegraphics[width=0.45\hsize]{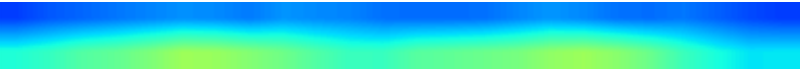}
        \end{tabular}
    }
    \\
    \subfigure[\textit{Residential} on depth ]
    {
        \begin{tabular}{cc}
            \includegraphics[width=0.45\hsize]{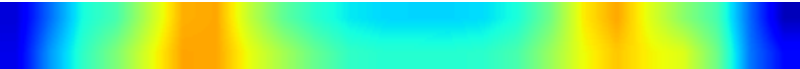}      \\
            \includegraphics[width=0.45\hsize]{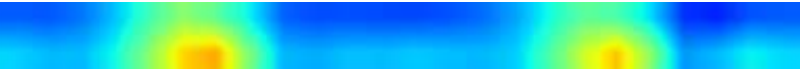}
        \end{tabular}
    }
    \hspace{5pt}
    \subfigure[\textit{Residential} on reflectance]
    {
        \begin{tabular}{cc}
            \includegraphics[width=0.45\hsize]{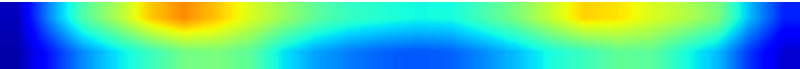}      \\
            \includegraphics[width=0.45\hsize]{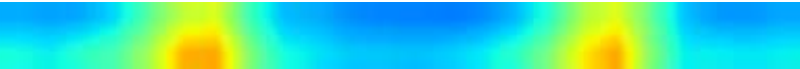}
        \end{tabular}
    }
    \\
    \subfigure[\textit{Urban} on depth ]
    {
        \begin{tabular}{cc}
            \includegraphics[width=0.45\hsize]{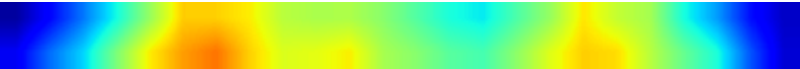}      \\
            \includegraphics[width=0.45\hsize]{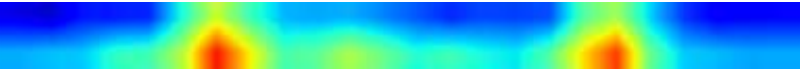}
        \end{tabular}
    }
    \hspace{5pt}
    \subfigure[\textit{Urban} on reflectance]
    {
        \begin{tabular}{cc}
            \includegraphics[width=0.45\hsize]{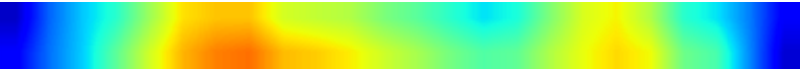}      \\
            \includegraphics[width=0.45\hsize]{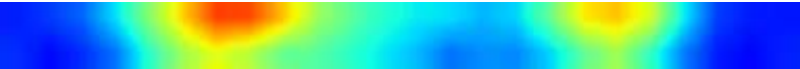}
        \end{tabular}
    }
    \caption{GradCAM of contributing regions which the CNN models activate to predict correct categories using Grad-CAM~\cite{Selvaraju_2017_ICCV}. Each group shows the average Grad-CAM maps on test sets obtained from the baseline VGG (top) and the modified VGG with RWMP and HCC (bottom). The modification works for sharpening of feature selectivity and equilibration of feature extractability between the center and the side area.}
    \label{fig:gradcam_all}
\end{figure}

Furthermore, we applied Grad-CAM to all given images on two types of models, the baseline VGG and VGG+RWMP+HCC, for depth and reflectance.
The averaged Grad-CAM maps are shown in Fig.~\ref{fig:gradcam_all}.
The categories are positioned up in rows and the modalities in columns.
Regardless of the modality, in \textit{Coast}, the center regions are strongly activated due to the distinctive appearance of the horizontal skyline (Fig.~\ref{fig:gradcam_all}(a)(b)).
In the three categories, \textit{ParkingIn}, \textit{Residential}, and \textit{Urban}, the regions of the front and back directions of the scanning vehicle are activated (Fig.~\ref{fig:gradcam_all}(e)(f)(i)--(l)).
These regions include skylines of artificial structures and roads.
In \textit{Forest} and the \textit{ParkingOut}, the maps show relatively uniform activations (Fig.~\ref{fig:gradcam_all}(c)(d)(g)(h)).
It can be seen that the contributing features appeared within local regions with a certain size, and everywhere on the image.
These ubiquitous features can be considered textural objects, such as woods, cars, or distant buildings.

Moreover, we can observe modality-independent characteristics in terms of the network modification.
Focusing particularly on \textit{Forest} and \textit{ParkingOut} again, we can see that the side regions are ignored on the baseline VGG, whereas the modified VGG responds to the regions that are intrinsically continuous but spatially isolated on the image.

These effects can also be seen in other categories, although slight differences occurred between the modalities.
For instance, in Fig.~\ref{fig:gradcam_all}(f) of a reflectance input, the network modification worked on equilibration of the feature extractability between the center and the side area.
These two areas correspond to the left and right sides of the scanning vehicle, respectively, and thus the appearances are reflected equally to the LiDAR.
On the other hand, in Fig.~\ref{fig:gradcam_all}(e) of a depth input, we can observe the imbalanced distribution on the areas.
This is mainly due to the geometrical difference while the vehicle drives along one side of the road.
It can be seen that the model learned the different features even if the same type of objects appeared there, and that leads to the difficulty of learning the model in comparison with the case of reflectance.

\subsection{Multi-modal Performance on Sparse MPO}
In this section, we report the performance of the multi-modal fusion approaches described in Section~\ref{sec:multimodal}.
For the Softmax Average and the Adaptive Fusion models, we used the models that showed the best results in the uni-modal experiments: the baseline VGG for the depth input and the VGG applying both HCC and RWMP for the reflectance input.
The results are shown in Table~\ref{table:multimodal}.

As shown in Table~\ref{table:multimodal}, all multi-modal models improved in accuracy in comparison to the uni-modal models except the Early Fusion model.
Particularly, the Softmax Average model shows the best results in the total score and the two categories, \textit{Forest}, and \textit{Urban}.
The Softmax Average was 0.69\% better than the best result in depth and 1.95\% better than the best result in reflectance.
The Adaptive Fusion model, which is similar to the Softmax Average model structurally, shows the best results in different two categories, we can see that the total accuracy is slightly inferior to the Softmax Average accuracy.
This is primarily due to the shortage of samples to train the gating network sufficiently.
As for the Early Fusion model, Long \textit{et al.}~\cite{long2015fully} observed a drop in accuracy when using the segmentation network with early-fused images. He suggested that the drop could be caused by the vanishing gradients, i.e. the difficulty to train the first layer that fuses modalities.
Likewise, our deep network can also cause this problem.
The Late Fusion model shows better results in a few categories in comparison to the uni-modal cases; however, the effectiveness is low.

In all fusing models, the two models that train separately and merge final scores have effective results. In contrast, the others that merge modalities within the networks have results equal to or less the results of the uni-modal models.
It can be seen that the efficiency of learning deep networks strongly affect the accuracies.

\begin{table*}[t]
    \tbl{Comparison of uni-modal and multi-modal models [\%]~(D: depth, R: reflectance)}
    {
        \centering
        \begin{tabular}{lllcccccccc}
            \toprule
            Models                   & Input &   & Coast          & Forest         & In. P.         & Out. P.        & Res.           & Urban          &   & Total          \\
            \midrule
            VGG                      & D     &   & 92.73          & 97.26          & \textbf{99.94} & 94.23          & 98.35          & 99.20          &   & 97.18          \\
            VGG + RWMP + HCC         & R     &   & 91.83          & 98.20          & 91.45          & \textbf{95.16} & 97.99          & 98.27          &   & 95.92          \\
            \cmidrule{1-2}
            \textbf{Softmax Average} & D + R &   & 94.27          & \textbf{98.38} & 99.58          & 94.91          & 99.12          & \textbf{99.56} &   & \textbf{97.87} \\
            Adaptive Fusion          & D + R &   & \textbf{94.59} & 98.20          & 99.77          & 94.85          & \textbf{99.19} & 99.37          &   & 97.62          \\
            Early Fusion             & D + R &   & 93.37          & 98.09          & 99.44          & 94.71          & 98.21          & 97.15          &   & 97.02          \\
            Late Fusion              & D + R &   & 93.49          & 97.57          & 99.25          & 94.23          & 98.39          & 98.88          &   & 97.19          \\
            \bottomrule
        \end{tabular}
    }
    \label{table:multimodal}
\end{table*}
\section{Conclusion}
\label{sec:conclusion}
In this paper we presented a new approach for outdoor place categorization by multi-modal convolutional neural networks.
We first constructed the large-scale outdoor place dataset named Multi-modal Panoramic 3D Outdoor (MPO), which contains a set of a paranoramic depth image, a panoramic reflectance image, and a corresponding place category.
This dataset contains two subsets in different resolutions; Sparse MPO and Dense MPO.

In the first experiment, we used either depth or reflectance inputs and trained four types of deep models based on the VGG~\cite{simonyan2014very}.
As a result, the baseline VGG showed the best result of 97.18\% for the depth input and the modified VGG showed the best result of 95.92\% for the reflectance input.
These results were better than the traditional hand-engineered features, Spin-Image, GIST and LBP.
In addition, we visualized the learned features in the networks and verified how they are activated for the specific categories.
In the second experiment, we used both of a depth and a reflectance images and trained the four types of multi-modal models.
The final results showed the best model was the Softmax Average, which merges the probabilities from the separately trained uni-modal models.

Future work includes increasing the number of categories in outdoor environments although it is difficult to clearly define the categories even when using human perception. Other future work is to design a novel model to fit the imbalance depth maps.
\section*{Acknowledgment}
This work was supported by JSPS KAKENHI Grant Number JP26249029.

\bibliographystyle{tADR}
\bibliography{main}

\begin{thebibliography}{10}
\providecommand{\url}[1]{\normalfont{#1}}
\providecommand{\urlprefix}{Available from: }
\providecommand{\eprint}[2][]{\url{#2}}

\bibitem{pronobis2012icra}
Pronobis A, Jensfelt P. Large-scale semantic mapping and reasoning with heterogeneous modalities. In: Proc. of the ieee int. conf. on robotics and automation (icra). Saint Paul, MN, USA. 2012 May. p. 3515--3522.

\bibitem{Stach2006}
Stachniss C, Mozos OM, Burgard W. Speeding up multi-robot exploration by considering semantic place information. In: Proc. of the ieee int. conf. on robotics and automation (icra). 2006. p. 1692--1697.

\bibitem{Koll2009}
Kollar T, Roy N. Utilizing object-object and object-scene context when planning to find things. In: Proc. of the ieee int. conf. on robotics and automation (icra). IEEE. 2009. p. 2168--2173.

\bibitem{Prono2010}
Pronobis A, Jensfelt P, Sj{\"o}{\"o} K, Zender H, Kruijff GJM, Mozos OM, Burgard W. Semantic modelling of space. In: Cognitive systems. Springer. 2010. p. 165--221.

\bibitem{Chris2010}
Christensen H, Kruijff GJM, Wyatt JL. Cognitive systems. Vol.~8. Springer Science \& Business Media. 2010.

\bibitem{ibanez2012autonomous}
Iba{\~n}ez-Guzman J, Laugier C, Yoder JD, Thrun S. Autonomous driving: Context and state-of-the-art. In: Handbook of intelligent vehicles. Springer. 2012. p. 1271--1310.

\bibitem{zhou2014learning}
Zhou B, Lapedriza A, Xiao J, Torralba A, Oliva A. Learning deep features for scene recognition using places database. In: Advances in neural information processing systems (nips). 2014. p. 487--495.

\bibitem{zhou2017places}
Zhou B, Lapedriza A, Khosla A, Oliva A, Torralba A. Places: A 10 million image database for scene recognition. IEEE Transactions on Pattern Analysis and Machine Intelligence. 2017;\hspace{0pt}.

\bibitem{Song:2015js}
Song S, Lichtenberg SP, Xiao J. {{SUN} {RGB-D}: A {RGB-D} scene understanding benchmark suite}. In: Proc. of the ieee conf. on computer vision and pattern recognition (cvpr). Boston, Massachusetts. 2015 June. p. 567--576.

\bibitem{Silberman:2012kh}
Silberman N, Hoiem D, Kohli P, Fergus R. {Indoor Segmentation and Support Inference from RGBD Images}. In: Proc. of the european conf. of computer vision (eccv). Florence, Italy. 2012 October. p. 746--760.

\bibitem{Mozos2012}
Mozos OM, Mizutani H, Kurazume R, Hasegawa T. Categorization of indoor places using the kinect sensor. Sensors. 2012;\hspace{0pt}12(5):6695--6711.

\bibitem{blanco2014ijrr}
Blanco-Claraco JL, Moreno-Due{\~n}as FA, Gonzalez-Jimenez J. The malaga urban dataset: High-rate stereo and lidar in a realistic urban scenario. The Int Journal of Robotics Research. 2014;\hspace{0pt}33:207--214.

\bibitem{fordcampus}
Pandey G, McBride JR, Eustice RM. Ford campus vision and lidar data set. The Int Journal of Robotics Research. 2011;\hspace{0pt}30:1543--1552.

\bibitem{geiger2012we}
Geiger A, Lenz P, Urtasun R. Are we ready for autonomous driving? the kitti vision benchmark suite. In: Proc. of the ieee conf. on computer vision and pattern recognition (cvpr). Providence, Rhode Island. 2012 June. p. 3354--3361.

\bibitem{torii201524}
Torii A, Arandjelovic R, Sivic J, Okutomi M, Pajdla T. 24/7 place recognition by view synthesis. In: Proc. of the ieee conf. on computer vision and pattern recognition (cvpr). 2015. p. 1808--1817.

\bibitem{lowe2004distinctive}
Lowe DG. Distinctive image features from scale-invariant keypoints. Int Journal of Computer Vision. 2004;\hspace{0pt}60(2):91--110.

\bibitem{xiao2010sun}
Xiao J, Hays J, Ehinger K, Oliva A, Torralba A, et~al.. Sun database: Large-scale scene recognition from abbey to zoo. In: Proc. of the ieee conf. on computer vision and pattern recognition (cvpr). San Francisco, CA. 2010 June. p. 3485--3492.

\bibitem{fazl2012ijrr}
Fazl-Ersi E, Tsotsos JK. Histogram of oriented uniform patterns for robust place recognition and categorization. The Int Journal of Robotics Research. 2012;\hspace{0pt}31(4):468--483.

\bibitem{Mozos2005}
Mozos OM, Stachniss C, Burgard W. Supervised learning of places from range data using adaboost. In: Proc. of the ieee int. conf. on robotics and automation (icra). IEEE. 2005. p. 1730--1735.

\bibitem{moral2013icra}
Fernandez-Moral E, Mayol-Cuevas W, Arévalo V, Gonzalez-Jimenez J. Fast place recognition with plane-based maps. In: Proc. of the ieee int. conf. on robotics and automation (icra). 2013 May. p. 2719--2724.

\bibitem{Mozos2013}
Mozos OM, Mizutani H, Jung H, Kurazume R, Hasegawa T. Categorization of indoor places by combining local binary pattern histograms of range and reflectance data from laser range finders. Advanced Robotics. 2013;\hspace{0pt}27(18):1455--1464.

\bibitem{arandjelovic2016netvlad}
Arandjelovic R, Gronat P, Torii A, Pajdla T, Sivic J. Net{VLAD}: {CNN} architecture for weakly supervised place recognition. In: Proc. of the ieee conf. on computer vision and pattern recognition (cvpr). 2016. p. 5297--5307.

\bibitem{gomez2015training}
Gomez-Ojeda R, Lopez-Antequera M, Petkov N, Gonzalez-Jimenez J. Training a convolutional neural network for appearance-invariant place recognition. arXiv preprint arXiv:150507428. 2015;\hspace{0pt}.

\bibitem{sunderhauf2015performance}
S{\"u}nderhauf N, Shirazi S, Dayoub F, Upcroft B, Milford M. On the performance of convnet features for place recognition. In: Proc. of the ieee/rsj int. conf. on intelligent robots and systems (iros). IEEE. 2015. p. 4297--4304.

\bibitem{UrsicICRA17}
Ur\u{s}i\u{c} P, Mandeljc R, Leonardis A, Kristan M. Part-based room categorization for household service robots. In: Proc. of the ieee int. conf. on robotics and automation (icra). 2016 May. p. 2287--2294.

\bibitem{maturana2015voxnet}
Maturana D, Scherer S. Voxnet: A 3d convolutional neural network for real-time object recognition. In: Proc. of the ieee/rsj int. conf. on intelligent robots and systems (iros). IEEE. 2015. p. 922--928.

\bibitem{wu20153d}
Wu Z, Song S, Khosla A, Yu F, Zhang L, Tang X, Xiao J. 3d shapenets: A deep representation for volumetric shapes. In: Proc. of the ieee conf. on computer vision and pattern recognition (cvpr). 2015. p. 1912--1920.

\bibitem{li2016vehicle}
Li B, Zhang T, Xia T. Vehicle detection from 3d lidar using fully convolutional network. arXiv preprint arXiv:160807916. 2016;\hspace{0pt}.

\bibitem{sizikova2016enhancing}
Sizikova E, Singh VK, Georgescu B, Halber M, Ma K, Chen T. Enhancing place recognition using joint intensity-depth analysis and synthetic data. In: Computer vision--eccv 2016 workshops. Springer. 2016. p. 901--908.

\bibitem{GoeddelIROS2016}
Goeddel R, Olson E. Learning semantic place labels from occupancy grids using cnns. In: Proc. of the ieee/rsj int. conf. on intelligent robots and systems (iros). 2016. p. 3999--4004.

\bibitem{simonyan2014very}
Simonyan K, Zisserman A. Very deep convolutional networks for large-scale image recognition. In: Proc. of the int. conf. on learning representations (iclr). 2015.

\bibitem{ioffe2015batch}
Ioffe S, Szegedy C. Batch normalization: Accelerating deep network training by reducing internal covariate shift. In: International conference on machine learning. 2015. p. 448--456.

\bibitem{he2016deep}
He K, Zhang X, Ren S, Sun J. Deep residual learning for image recognition. In: Proc. of the ieee conf. on computer vision and pattern recognition (cvpr). 2016. p. 770--778.

\bibitem{shi2015deeppano}
Shi B, Bai S, Zhou Z, Bai X. Deeppano: Deep panoramic representation for 3-d shape recognition. IEEE Signal Processing Letters. 2015;\hspace{0pt}22(12):2339--2343.

\bibitem{mees2016choosing}
Mees O, Eitel A, Burgard W. Choosing smartly: Adaptive multimodal fusion for object detection in changing environments. In: Proc. of the ieee/rsj int. conf. on intelligent robots and systems (iros). IEEE. 2016. p. 151--156.

\bibitem{moody1995simple}
Moody J, Hanson S, Krogh A, Hertz JA. A simple weight decay can improve generalization. Advances in Neural Information Processing Systems (NIPS). 1995;\hspace{0pt}4:950--957.

\bibitem{paszke2017automatic}
Paszke A, Gross S, Chintala S, Chanan G, Yang E, DeVito Z, Lin Z, Desmaison A, Antiga L, Lerer A. Automatic differentiation in pytorch. In: Nips-w. 2017.

\bibitem{John1997}
Johnson AE. Spin-images: A representation for 3-d surface matching. [Ph.D. thesis]. The Robotics Institute, Carnegie Mellon University. 1997.

\bibitem{oliva2001modeling}
Oliva A, Torralba A. Modeling the shape of the scene: A holistic representation of the spatial envelope. Int Journal of Computer Vision. 2001;\hspace{0pt}42(3):145--175.

\bibitem{Ojala2002}
T~Ojala MP, M\"{a}enp\"{a}\"{a} T. Multiresolution gray-scale and rotation invariant texture classification with local binary patterns. IEEE Trans on Pattern Analysis and Machine Intelligence. 2002;\hspace{0pt}24(7):971--987.

\bibitem{Selvaraju_2017_ICCV}
Selvaraju RR, Cogswell M, Das A, Vedantam R, Parikh D, Batra D. Grad-cam: Visual explanations from deep networks via gradient-based localization. In: The ieee international conference on computer vision (iccv). 2017 Oct.

\bibitem{long2015fully}
Long J, Shelhamer E, Darrell T. Fully convolutional networks for semantic segmentation. In: Proc. of the ieee conf. on computer vision and pattern recognition (cvpr). 2015. p. 3431--3440.

\end{thebibliography}

\appendix
\section{Specifications of sensors}

\begin{table}[htb]
    \tbl{Specifications of Garmin GPS 18x LVC}
    {
        \centering
        \begin{tabularx}{0.7\hsize}{LL}
            \toprule
            Receiver Sensitivity & $-185$ dBW minimum   \\ \midrule
            Acquisition Times    & 1 s ~                \\ \midrule
            Accuracy (Standard)  & 15 m, RMS 95\% typ   \\ \midrule
            Accuracy (DGPS)      & 3--5 m, RMS 95\% typ \\ \midrule
            Accuracy (WAAS)      & $<$3 m, RMS 95\% typ \\ \midrule
            Velocity             & 0.1 m/s RMS (Const.) \\ \midrule
            Interval             & 1PPS (20--980 ms)    \\
            \bottomrule
        \end{tabularx}
    }
    \label{table:gps}
\end{table}

\begin{table}[htb]
    \tbl{Specifications of Velodyne HDL-32e}
    {
        \centering
        \begin{tabularx}{0.7\hsize}{llL}
            \toprule
            \multicolumn{2}{l}{Channels}          & 32 Channels ToF                                                                    \\  \midrule
            \multicolumn{2}{l}{Measurement Range} & 1 $\sim$ 70m\ Up to 100 m                                                          \\ \midrule
            \multicolumn{2}{l}{Accuracy}          & $\pm$ 2 cm (1 $\sigma@25m$)                                                        \\  \midrule
            Field of View      & Vertical   & +10.67$^\circ$ $\sim$ -30.67$^\circ$ (41.33$^\circ$) \\ \cmidrule{2-2}
                               & Horizontal & 360$^\circ$                                          \\  \midrule
            Angular Resolution & Vertical   & 1.33$^\circ$                                         \\  \cmidrule{2-2}
                               & Horizontal & 0.1$^\circ$ $\sim$ 0.4$^\circ$                       \\  \midrule
            \multicolumn{2}{l}{Rotation Rate}     & 5 Hz  $\sim$  20 Hz                                                                \\ \midrule
            \multicolumn{2}{l}{Class}             & Class 1 Eye Safe                                                                   \\ \midrule
            \multicolumn{2}{l}{Wavelengnth}       & 905nm                                                                              \\
            \bottomrule
        \end{tabularx}
    }
    \label{table:velodyne}
\end{table}

\begin{table}[htb]
    \tbl{Specifications of FARO Focus 3D S120}
    {
        \centering
        \begin{tabularx}{0.7\hsize}{llL}
            \toprule
            \multicolumn{2}{l}{Measurement Range} & 0.6 $\sim$ 120m                 \\ \midrule
            \multicolumn{2}{l}{Accuracy}          & $\pm$ 2 mm                      \\  \midrule
            Field of View      & Vertical   & 300$^\circ$   \\ \cmidrule{2-3}
                               & Horizontal & 360$^\circ$   \\  \midrule
            Angular Resolution & Vertical   & 0.009$^\circ$ \\ \cmidrule{2-3}
                               & Horizontal & 0.009$^\circ$ \\  \midrule
            \multicolumn{2}{l}{Class}             & Class 3R (20mW)                 \\ \midrule
            \multicolumn{2}{l}{Wavelength}        & 905nm                           \\
            \bottomrule
        \end{tabularx}
    }
    \label{table:FARO}
\end{table}

\end{document}